\documentclass{article}

\PassOptionsToPackage{numbers,compress,sort}{natbib}
\usepackage[final]{neurips_2022}

\usepackage[utf8]{inputenc} 
\usepackage[T1]{fontenc}    
\usepackage{hyperref}       
\usepackage{url}            
\usepackage{booktabs}       
\usepackage{amsfonts}       
\usepackage{nicefrac}       
\usepackage{microtype}      
\usepackage{xcolor}         

\usepackage{adjustbox}
\usepackage{subfigure}
\usepackage{mathtools}
\usepackage{amsthm}
\usepackage{xcolor}
\usepackage{amssymb}
\usepackage{algorithm}
\usepackage{algorithmic}
\usepackage{thm-restate}
\theoremstyle{plain}
\newtheorem{theorem}{Theorem}[section]
\newtheorem{proposition}[theorem]{Proposition}
\newtheorem{lemma}[theorem]{Lemma}

\theoremstyle{definition}

\newtheorem{remark}[theorem]{Remark}

\DeclareMathOperator*{\argmax}{arg\,max}
\DeclareMathOperator*{\argmin}{arg\,min}

\newcommand{\E}{\mathbb{E}}
\newcommand{\bE}{\mathbf{\hat{E}}}

\newcommand{\R}{\mathbb{R}}

\newcommand{\KL}{D_{\mathrm{KL}}}

\newcommand{\redb}[1]{\textcolor{red}{\bar{#1}}}
\newcommand{\red}[1]{\textcolor{red}{#1}}
\newcommand{\blue}[1]{\textcolor{blue}{#1}}
\newcommand{\revised}[1]{#1}

\def\gB{{\mathcal{B}}}

\def\gL{{\mathcal{L}}}

\def\gT{{\mathcal{T}}}

\title{LobsDICE: Offline Learning from Observation via Stationary Distribution Correction Estimation}

\author{%
  Geon-Hyeong Kim$^{1,3,*,\dagger}$,
  Jongmin Lee$^{4,*}$,  
  Youngsoo Jang$^{1,3,\dagger}$,\\
  \textbf{
  Hongseok Yang$^{1,2,5}$   
  Kee-Eung Kim$^{1,2}$  
  }\\
  $^1$ School of Computing, KAIST
  \\
  $^2$ Kim Jaechul Graduate School of AI, KAIST
  \\
  $^3$ LG AI Research
  \\
  $^4$ University of California, Berkeley\\
  $^5$ Discrete Mathematics Group, Institute for Basic Science (IBS)
}

\begin{document}

\maketitle
\def\thefootnote{$*$}\footnotetext{Equal contribution.} 
\def\thefootnote{$\dagger$}\footnotetext{Work done while the authors were students at KAIST.}
\begin{abstract}
We consider the problem of learning from observation (LfO), in which the agent aims to mimic the expert's behavior from the state-only demonstrations by experts. We additionally assume that the agent cannot interact with the environment but has access to the action-labeled transition data collected by some agents with unknown qualities. This offline setting for LfO is appealing in many real-world scenarios where the ground-truth expert actions are inaccessible and the arbitrary environment interactions are costly or risky. In this paper, we present LobsDICE, an offline LfO algorithm that learns to imitate the expert policy via optimization in the space of stationary distributions. Our algorithm solves a single convex minimization problem, which minimizes the divergence between the two state-transition distributions induced by the expert and the agent policy. Through an extensive set of offline LfO tasks, we show that LobsDICE outperforms strong baseline methods.
\end{abstract}

\section{Introduction}
\label{sec:intro}
The ability to learn from experience is one of the core aspects of an intelligent agent. Reinforcement learning (RL)~\citep{sutton1998rlbook} provides a framework to acquire such an  intelligent behavior autonomously through interactions with the environment while receiving reward feedback.
However, the practical applicability of RL to real-world domains has been limited for two reasons.
First, designing a suitable reward function for complex tasks can be extremely difficult. 
The RL agent learns behaviors incentivized by the reward function rather than the ones intended, which can be nontrivial to specify in terms of reward. Second, the need for online interaction with the environment during the RL training loop has hindered its adoption in many real-world domains, where environment interactions are costly or risky.

Imitation learning (IL)~\citep{schaal1997lfd,argall2009survey,ross2011reduction} circumvents the difficulty of reward design in RL by leveraging demonstrations given by experts, where the goal is to mimic the expert's behavior.
However, the standard IL requires the expert demonstrations to contain not only the state information (e.g. robot joint angles) but also the precise action (e.g. robot joint torques) executed by the expert at each time step.
This demand for explicit action labels is in contrast to the way human imitates (e.g. learning by watching videos) and precludes leveraging a massive amount of data in which the action label is missing.
Therefore, developing an imitation learning algorithm that can learn from observing the experts' state-only trajectories is a promising direction for creating a practical, autonomous intelligent agent.
Learning from observation (LfO)~\citep{Bentivegna2004robot,liu2018ifovideo,torabi2018gaifo,ZhuEtal2020opolo,kidambi2021mobile} concerns this particular learning scenario and has been gaining interest in recent years.

In this paper, we are particularly interested in solving the LfO problem in an \emph{offline} setting: Given the state-only demonstrations by experts and abundant state-action demonstrations by imperfect agents of arbitrary levels of optimality, we aim to find a policy that follows the sequence of expert's states without further interaction with the environment.
This problem setting is appealing in many practical situations where environment interactions are costly or risky.
Yet, it cannot be  straightforwardly tackled by existing approaches. Most of the existing LfO methods are on-policy algorithms~\citep{torabi2018gaifo,yang2019iddm,sun2019fail,liu2019sail}, which is not directly applicable to the offline setting.
One of the few exceptions is BCO~\citep{torabi2018bco}, which performs behavior cloning on the inferred action via an inverse dynamics model.
This method could be suboptimal unless the inverse dynamics disagreement is always zero, i.e. the underlying environment dynamics is deterministic and injective.
Another one is OPOLO~\citep{ZhuEtal2020opolo}, an off-policy LfO algorithm, but it relies on nested min-max optimization as well as out-of-distribution (OOD) action values, 
which can be unstable especially in the offline setting.
IQ-Learn~\citep{garg2021iq} solves an Inverse RL and can be applied to the offline LfO by learning a state-only reward function. However, it suffers from numerical instability due to using OOD action values in the offline setting.
Lastly, RCE~\citep{eysenbach2021replacing}, an example-based control algorithm, can in principle be applied to the offline LfO setting by providing the expert trajectories as successful example states; however, its empirical performance is known to be limited without online data collection.

We present an offline LfO algorithm that minimizes the divergence between state-transition distributions induced by the expert and the learned policy, 
without requiring an inverse dynamics model.
Our algorithm, \textit{offline Learning from OBServation via stationary DIstribution Correction Estimation} (\mbox{LobsDICE}),
essentially optimizes in the space of state-action stationary distributions and state-transition stationary distributions, rather than in the space of policies.
We show that our formulation can be reduced to a single convex minimization problem that can be solved efficiently in practice, unlike existing imitation learning (from observation) algorithms that rely on nested min-max optimization.
In the experiments, we demonstrate that LobsDICE can successfully recover the state visitations by the expert, outperforming strong baseline methods. 

\section{Preliminaries}
\subsection{Markov decision process}
We consider an environment modeled as a Markov Decision Process (MDP), defined by $M=\langle S,A,T,R,p_0,\gamma\rangle$~\citep{sutton1998rlbook}, where $S$ is the set of states, $A$ is the set of actions, $T:S\times A\rightarrow\Delta(S)$ is the transition probability, $R:S\times A\rightarrow\R$ is the reward function, $p_0\in\Delta(S)$ is the distribution of the initial state, and $\gamma\in(0,1)$ is the discount factor.
The policy $\pi:S\rightarrow\Delta(A)$ is a mapping from states to distribution over actions.
For the given policy $\pi$, its \textit{state-action} stationary distribution $d^\pi(s,a)$ and \textit{state-transition} stationary $\bar d^\pi(s,s')$ are defined as:
\begin{align*}
    d^\pi(s,a) &:= (1-\gamma)\textstyle\sum\limits_{t=0}^\infty\gamma^t\Pr(s_t=s,a_t=a),~~~ 
    \bar d^\pi(s,s') := (1-\gamma)\textstyle\sum\limits_{t=0}^\infty\gamma^t\Pr(s_t=s,s_{t+1}=s'),
\end{align*}
where $s_0\sim p_0$, $a_t\sim\pi(\cdot|s_t)$, and $s_{t+1}\sim T(\cdot|s_t,a_t)$ for all timesteps $t\ge 0$.
For brevity, the bar notation $\bar{(\cdot)}$ will be used to denote the distributions for $(s,s')$, e.g. $\bar d^\pi(s,s')$.

We assume offline LfO setting, where direct, online interactions with the environment are not allowed, and the policy should be optimized solely from a pre-collected dataset. We denote the dataset of state-only demonstrations collected by experts as $D^E = \{(s,s')_{i}\}_{i=1}^{N_E}$ and the dataset of state-action demonstrations by some imperfect agents as $D^I = \{(s,a,s')_i\}_{i=1}^{N_I}$.
That is, we do not have information about the actions taken by the expert, but instead, we have additional action-labeled transition data collected by some other agents with unknown levels of optimality.
We denote the corresponding distributions of the datasets $D^E$ and $D^I$ by $\bar{d}^E$ and $d^I$, respectively.
For brevity, we will abuse notation $d^I$ to represent $(s,a)\sim d^I$, $(s,a,s')\sim d^I$, and $(s,s')\sim \bar{d}^I$ unless ambiguous.

\subsection{Imitation learning and learning from observation}
\textit{Imitation learning} (IL) aims to mimic the expert policy from its state-action demonstrations.
IL can be naturally formulated as a distribution matching problem that minimizes the divergence between \emph{state-action} stationary distributions induced by the expert and the target policy~\citep{ho2016gail,ke2020generalil}.
For example, one can consider minimizing KL-divergence~\citep{kostrikov2020valuedice}:
\begin{align}
    \min_\pi\KL(d^\pi(s,a)\|d^E(s,a))=\E_{(s,a) \sim d^\pi} \Big[\log\tfrac{d^\pi(s,a)}{d^E(s,a)}\Big].
    \label{eq:kl_divergence_state_action}
\end{align}

However, the standard IL requires action labels in the expert demonstrations, which may be a too strong requirement for various practical situations.
\textit{Learning from observation} (LfO) relaxes the requirement 
on action labels, and aims to imitate the expert's behavior only from the state observations.
Since the expert's action information is missing in the demonstrations, the distribution matching for state-action stationary distribution is no longer readily applicable.
Therefore, LfO is reformulated as another distribution matching problem that minimizes the divergence between \emph{state-transition} stationary distributions induced by the expert and the target policy~\citep{yang2019iddm,torabi2018gaifo,ZhuEtal2020opolo}\footnote{See Appendix~\ref{app:why_ss_matching} for a discussion of why $\bar d^\pi(s,s')$-matching is preferable to $\bar d^\pi(s)$-matching.}:
\begin{align}
    \min_\pi\KL(\bar{d}^\pi(s,s')\|\bar{d}^E(s,s'))=\E_{(s,s') \sim \bar{d}^\pi}\Big[\log\tfrac{\bar{d}^\pi(s,s')}{\bar{d}^E(s,s')}\Big].
    \label{eq:kl_divergence_state_transition}
\end{align}
Still, optimizing Eq.~\eqref{eq:kl_divergence_state_transition} in a purely offline manner is challenging since naively estimating the 
expectation 
would require the knowledge of $T(\cdot | s, a)$ for the OOD action $a \sim \pi(s)$ by the target policy, which is inaccessible in the offline LfO setting; see Section 9.8 in \citep{ZhuEtal2020opolo} for more discussions.
OPOLO~\citep{ZhuEtal2020opolo}, an off-policy LfO algorithm, mitigates this challenge by minimizing the following \emph{upper bound} of the divergence $\KL(\bar d^\pi(s,s') \| \bar d^E(s,s'))$:
\begin{align}
    \eqref{eq:kl_divergence_state_transition}
    \le \E_{\bar{d}^\pi(s,s')} \Big[ \log \tfrac{ \bar d^I(s,s') }{\bar d^E(s,s')} \Big] + \KL( d^\pi(s,a) \| d^I(s,a) )
    \label{eq:opolo_upper_bound}
\end{align}
and applying DualDICE trick~\citep{nachum2019dualdice} to the RHS of \eqref{eq:opolo_upper_bound}. The upper bound gap is given by 
the inverse dynamics disagreement between the target policy and the imperfect demonstrator:
\begin{align}
    \eqref{eq:opolo_upper_bound} - \eqref{eq:kl_divergence_state_transition} = \KL(d^\pi(a | s,s') \| d^I(a | s, s')) ,
    \label{eq:opolo_gap}
\end{align}
which usually gets larger as the stochasticity of the environment transition increases. 

\section{LobsDICE}
\label{sec:lobsdice}
We present \textit{offline Learning from OBServation via stationary DIstribution Correction Estimation} (LobsDICE), a principled offline LfO algorithm that further extends the recent progress made by
the DIstribution Correction Estimation (DICE) methods for offline RL.
LobsDICE essentially optimizes the stationary distributions of the target policy to match the expert's state visitations.

In the context of offline RL, the policy constraint principle (i.e. prevent deviating too much from the data support) is one of the common approaches to 
avoid severe performance degradation~\citep{nachumE2019algaedice,fujimoto2019bcq,kumar2019bear,lee2021optidice,jin2021pessimism}.
In the same manner, we use KL divergence minimization between $\bar{d}^\pi(s,s')$ and $\bar{d}^E(s,s')$ with additional KL regularization on
the deviation from $d^I$:
\begin{equation}
    \min_\pi\KL(\red{\bar{d}^\pi(s,s')} \| \bar{d}^E(s,s') ) + \alpha\KL( \blue{d^\pi(s,a)} \| d^I(s,a)).
    \label{eq:state_transition_matching}
\end{equation}
Here the hyperparameter $\alpha>0$ balances between encouraging state-transition matching and preventing distribution shift from the distribution of imperfect demonstrations.
All the proofs can be found in Appendix~\ref{app:theoretical_analysis}.

\subsection{Lagrange dual formulation}
The derivation of our algorithm starts by rewriting the (regularized) distribution matching problem~\eqref{eq:state_transition_matching} in terms of directly optimizing stationary distribution, rather than policy:
\begin{align}
    \max_{\blue{d},\redb{d}\ge0}~ 
    &-\KL(\redb{d}(s,s') \| \bar{d}^E(s,s') ) - \alpha\KL(\blue{d}(s,a) \| d^I(s,a)) \label{eq:main_objective}\\
    \text{s.t.}~
    &\textstyle\sum\limits_{a'} \blue{d}(s',a') = (1-\gamma)p_0(s') + \gamma \sum\limits_{s,a} \blue{d}(s,a) T(s'|s,a) \quad\forall s',\label{eq:bellman_flow_constraint}\\
    &\textstyle \sum\limits_{a}\blue{d}(s,a) T(s'|s,a) = \redb{d}(s,s')\quad\forall s,s',\label{eq:marginalization_constraint}
\end{align}
The Bellman flow constraint~\eqref{eq:bellman_flow_constraint} ensures $d(s,a)$ to be a valid state-action stationary distribution of some policy, where $d(s,a)$ can be interpreted as a normalized occupancy measure of $(s,a)$.
The marginalization constraint~\eqref{eq:marginalization_constraint} enforces $\bar d(s,s')$ to be the state-transition stationary distribution that is directly induced by $d(s,a)$.
In essence, the constrained optimization problem (\ref{eq:main_objective}-\ref{eq:marginalization_constraint}) seeks the stationary distributions of an optimal policy, which best matches the state-transition trajectories of the expert.
Once we have computed the optimal solution $(d^*, \bar d^*)$, its corresponding optimal policy can also be obtained by normalizing $d^*$ for each state~\cite{puterman1994markov}:
$\pi^*(a|s)=\frac{d^*(s,a)}{\sum_a d^*(s,a)}$.

Note that DemoDICE~\citep{kim2022demodice} considers a similar optimization problem to ours, but it deals with the offline IL (i.e. \emph{state-action} stationary distribution matching), whereas we consider the offline LfO (e.g \emph{state-transition} stationary distribution matching).
Accordingly, the optimization variable $\redb{d}(s,s')$ and the marginalization constraint~\eqref{eq:marginalization_constraint} are newly added in our formulation.

We then consider the Lagrangian for the constrained optimization (\ref{eq:main_objective}-\ref{eq:marginalization_constraint}):
\begin{align}
    \min_{\mu,\nu} \max_{\blue{d},\redb{d}\ge0}
    &- \E_{\redb{d}}\Big[ \log \tfrac{\redb{d}(s,s')}{\bar d^E(s,s')} \Big] - \alpha \E_{\blue{d}}\Big[ \log \tfrac{\blue{d}(s,a)}{d^I(s,a)} \Big]
    + \textstyle\sum\limits_{s,s'} \mu(s,s') \big( \redb{d}(s,s') - \sum\limits_{a} \blue{d}(s,a) T(s'|s,a) \big) \nonumber \\
    & + \textstyle\sum\limits_{s'} \nu(s') \big( (1-\gamma)p_0(s') + \gamma \sum\limits_{s,a} \blue{d} (s,a) T(s'|s,a) - \sum\limits_{a'}\blue{d}(s',a') \big),
    \label{eq:lagrangian}
\end{align}
where $\nu(s) \in \R$ are the Lagrange multipliers for the Bellman flow constraints~\eqref{eq:bellman_flow_constraint}, and $\mu(s,s') \in \R$ are the the Lagrange multipliers for the marginalization constraint~\eqref{eq:marginalization_constraint}.
Note that the Lagrangian \eqref{eq:lagrangian} cannot be naively optimized in an offline manner since it requires evaluation of $T(s'|s,a)$ for $(s,a) \sim d$, which is not accessible in the offline LfO setting.
Therefore, we rearrange the terms in \eqref{eq:lagrangian} to eliminate the direct dependence on $d$ and $\bar{d}$, introducing new optimization variables $w$ and $\bar w$ that denote stationary distribution correction ratios for $(s,a)$ and $(s,s')$, respectively:
\begin{align}
    \eqref{eq:lagrangian}=&\min_{\mu,\nu} \max_{\blue{d},\redb{d}\ge0} (1 - \gamma) \E_{s_0 \sim p_0} [\nu(s_0)]  
    + \E_{(s,s') \sim \redb{d}} \Big[ \mu(s,s') 
    \overbrace{- \log\underbrace{\tfrac{\redb{d}(s,s')}{\bar{d}^I(s,s')}}_{=:\redb{w}(s,s')} + \underbrace{\log\tfrac{\bar{d}^E(s,s')}{\bar{d}^I(s,s')}}_{=: r(s,s')}}^{= -\log \frac{\redb{d}(s,s')}{ d^E(s,s') }} \Big] \nonumber \\
    & \hspace{30pt} + \E_{(s,a) \sim \blue{d}} \Big[ \underbrace{\E_{s'}[-\mu(s,s') + \gamma \nu(s')] - \nu(s)}_{=:e_{\mu,\nu}(s,a)} - \alpha\log\underbrace{\tfrac{\blue{d}(s,a)}{d^I(s,a)}}_{=:\blue{w}(s,a)} \Big] \label{eq:using_transpose_operator} \\
    =& \min_{\mu,\nu} \max_{\blue{w},\redb{w}\ge0} (1 - \gamma) \E_{s_0 \sim p_0} [\nu(s_0)] + \E_{(s,s') \sim \bar{d}^I} \big[ \redb{w}(s,s') \big( r(s,s') + \mu(s,s') - \log\redb{w}(s,s') \big)\big] \nonumber \\
    & \hspace{30pt} + \E_{(s,a) \sim d^I} \big[\blue{w}(s,a) \big( e_{\mu,\nu}(s,a) - \alpha\log\blue{w}(s,a) \big)\big] =: \gL(\blue{w},\redb{w},\mu,\nu). \label{eq:maximin_lagrangian_detail}
\end{align}
Similar to the assumption of full coverage which is fairly standard across a broad set of recent offline RL approaches~\citep{kumar2020cql, lee2021optidice, ma2021conservative}, we make a milder assumption that $\bar{d}^I(s,s')>0$ whenever $\bar{d}^E(s,s')>0$.
This assumption is necessary to recover the expert's behavior successfully.
We introduce the log ratio $r(s,s') = \log \frac{\bar d^E(s,s')}{\bar d^I(s,s')}$ in \eqref{eq:using_transpose_operator} to take an expectation under $\bar d^I$ (instead of $\bar d^E$), which is assumed to have a broader support than $\bar d^E$.
This log ratio $r(s,s')$ can be easily estimated using a pretrained discriminator for two datasets $D^E$ and $D^I$, which will be explained in detail in the following section.

In summary, LobsDICE aims to solve the minimax optimization,
\begin{equation}
    \min_{\mu,\nu} \max_{\blue{w},\redb{w} \ge 0} \gL(\blue{w},\redb{w},\mu,\nu).
    \label{eq:maximin_lagrangian}
\end{equation}
The optimal solution $(\blue{w}^*, \redb{w}^*)$ of \eqref{eq:maximin_lagrangian} represents stationary distribution corrections of an optimal policy $\pi^*$: $\blue{w}^*(s,a)=\frac{\blue{d}^{\pi^*}(s,a)}{d^I(s,a)}$ and $\redb{w}^*(s,s')=\frac{\redb{d}^{\pi^*}(s,s')}{\bar{d}^I(s,s')}$.

\subsection{Log ratio estimation via a pretrained discriminator}
To optimize \eqref{eq:maximin_lagrangian_detail}, an estimate of the log ratio $r(s,s') = \log\frac{\bar{d}^E(s,s')}{\bar{d}^I(s,s')}$ is required.
The log ratio estimation is straightforward for tabular MDPs since we 
can use empirical distributions from the datasets to estimate 
$\bar d^E(s,s')$ and $\bar d^I(s,s')$.
For continuous MDPs, we train a discriminator $c:S\times S\rightarrow[0,1]$ by solving the following maximization problem~\citep{goodfellow2014generative,ZhuEtal2020opolo}:
\begin{align}
    c^* = \hspace{-5pt} \argmax_{c:S\times S\rightarrow[0,1]}
    \E_{(s,s') \sim \bar{d}^E}[\log c(s,s')] + \E_{(s,s') \sim \bar{d}^I}[\log (1-c(s,s'))].
    \label{eq:discriminator_training}
\end{align}
It is easy to show that the optimal discriminator satisfies $c^*(s,s')=\frac{\bar{d}^E(s,s')}{\bar{d}^E(s,s')+\bar{d}^I(s,s')}$.
Thus, $r(s,s')$ can be derived from the optimal discriminator $c^*$ as,
\begin{equation}
    r(s,s') = -\log\big( \tfrac{1}{c^*(s,s')} - 1 \big).
\end{equation}

\subsection{Minimax to min: a closed-form solution}
Exploiting the strict convexity of $x \log x$, we can derive a closed-form solution to the inner maximization for $(w, \bar w)$ in \eqref{eq:maximin_lagrangian_detail}.
\begin{restatable}{proposition}{firstclosedformsolution}
\label{prop:closed_form_solution}
For any $(\mu, \nu)$, the closed-form solution to the inner maximization of \eqref{eq:maximin_lagrangian_detail}, i.e. $(\blue{w}_{\mu,\nu}, \redb{w}_{\mu}) = \argmax_{\blue{w},\redb{w} \ge 0} \gL (\blue{w},\redb{w},\mu,\nu)$, is given by:
\begin{align}
    \blue{w}_{\mu,\nu}(s,a)&=\exp\big(\tfrac{1}{\alpha} e_{\mu,\nu}(s,a) -1\big) 
    \emph{ ~and~ }
    \redb{w}_{\mu}(s,s')=\exp(r(s,s')+\mu(s,s')-1). \label{eq:closed_form_solution_wsa_wss1}
\end{align}
\end{restatable}
Based on the above result, we reduce the nested min-max optimization of \eqref{eq:maximin_lagrangian_detail} to a single minimization by plugging the closed-form solution ($\blue{w}_{\mu,\nu}, \redb{w}_{\mu}$) into $\gL(\blue{w},\redb{w},\mu,\nu)$:
\begin{align}
    \min_{\mu,\nu}\, & \gL (\blue{w}_{\mu,\nu},\redb{w}_{\mu},\mu, \nu) 
    = (1-\gamma)\E_{s \sim p_0}[\nu(s)] + \E_{(s,s') \sim \bar{d}^I} \big[ \exp\big(r(s,s')+\mu(s,s')-1\big) \big] \nonumber \\
    &  + \alpha\E_{(s,a) \sim d^I}\big[ \exp\big(\tfrac{1}{\alpha} e_{\mu,\nu}(s,a) -1\big)\big]. \label{eq:lagrange_objective_double}
\end{align}
We can even show that $\gL (\blue{w}_{\mu,\nu},\redb{w}_{\mu},\mu, \nu)$ is a convex function of $\mu$ and $\nu$.
\begin{restatable}{proposition}{convexitydouble}
$\gL (\blue{w}_{\mu,\nu},\redb{w}_{\mu},\mu, \nu)$ is convex with respect to $\mu$ and $\nu$.
\label{prop:convexity_double}
\end{restatable}

In short, by operating in the space of stationary distributions, offline LfO can, in principle, be resolved by solving a \emph{convex minimization} problem.
This is in contrast to the existing LfO algorithms, which typically involve either an adversarial training that optimizes the policy and the discriminator~\citep{torabi2018gaifo,yang2019iddm,kidambi2021mobile} or a nested min-max optimization for the policy and the critic~\citep{ZhuEtal2020opolo}.

\subsection{Policy extraction}
So far, we have derived an algorithm that essentially solves the state-transition distribution matching problem via convex minimization.
However, we obtain $(\mu^*, \nu^*)$ as the solution of \eqref{eq:lagrange_objective_double}, instead of the optimal policy $\pi^*$.
The remaining problem is to extract the optimal policy from $(\mu^*, \nu^*)$. 
The first step is to see that we can obtain the state-action stationary distribution correction $\blue{w}_{\mu^*, \nu^*}(s,a) = \frac{\blue{d}^{\pi^*}(s,a)}{d^I(s,a)}$ of the optimal policy from the closed-form solution~\eqref{eq:closed_form_solution_wsa_wss1}.
Among many possibilities to extract the policy from the state-action distribution correction, we adopt weighted behavior cloning (WBC):
\begin{align}
    &\max_{\pi} \E_{(s,a) \sim d^{\pi^*}}[ \log \pi(a|s) ] = \E_{(s,a) \sim d^I} [ \blue{w}_{\mu^*, \nu^*}(s,a) \log \pi(a|s) ] 
    \label{eq:policy_extraction_wbc}
\end{align}
which aims to maximize the predicted probability of actions chosen by optimal policy $\pi^*$. 
This is done by BC on the offline dataset $D^I$ where each sample $(s,a)$ is weighted by $\blue{w}_{\mu^*,\nu^*}(s,a)$.
For tabular MDPs, we can formally show that WBC extracts an optimal policy $\pi^*$ (Appendix~\ref{app:policy_extraction_tabular}).

\subsection{Practical algorithm with sample-based approximation}
\label{subsec:pratical_algorithm}
In practice, 
we estimate $\gL(\blue{w},\redb{w},\mu,\nu)$ in \eqref{eq:maximin_lagrangian_detail} using samples from distribution $d^I$. 
Let $\bE_{x \in D}[ f(x) ] := \frac{1}{|D|} \sum_{x\in D} f(x)$ be a Monte-Carlo estimate of $\E_{x \sim p}[f(x)]$ where $D = \{x_i\}_{i=1}^{|D|} \sim p$.
We denote each sample $(s,a,s')$ in $D^I$ as $x$ for brevity.
\begin{align}
    &\min_{\mu, \nu} \max_{\blue{w}, \redb{w} \ge 0} \widehat \gL(\blue{w},\redb{w},\mu,\nu) 
    :=  (1 - \gamma) \bE_{s_0 \in D_0} [\nu(s_0)] \label{eq:maximin_lagrangian_detail_sample} \\
    &~~~+ \bE_{x \in D^I} \big[ \redb{w}(s,s') \big( r(s,s') + \mu(s,s') - \log\redb{w}(s,s') \big)  + \blue{w}(s,a) \big( \hat e_{\mu,\nu}(s,a,s') - \alpha\log\blue{w}(s,a) \big)\big] \nonumber
\end{align}
where $\hat{e}_{\mu, \nu}(s,a,s'):=-\mu(s,s')+\gamma\nu(s')-\nu(s)$ is a single-sample estimate of $e_{\mu, \nu}(s,a)$.
Note that this sample-based objective function $\widehat \gL(w, \bar w, \mu,\nu)$ can be estimated only from samples in the offline dataset $D^I$ and is an unbiased estimator of $\gL(w, \bar w, \mu, \nu)$ as long as every sample in $D^I$ was collected by interacting with the underlying MDP.
To the best of our knowledge, this is the first result to directly solve the state-transition distribution matching problem in a fully offline manner. In contrast, OPOLO~\citep{ZhuEtal2020opolo} relies on the (potentially loose) \emph{upper bound} of the objective in \eqref{eq:opolo_upper_bound}.

To reduce the minimax optimization to a single minimization, we apply the non-parametric closed-form solution for each sample $x = (s,a,s')$ in $D^I$:
\begin{align}
    \blue{\widehat{w}}_{\mu,\nu}(x) = \exp\Big(\tfrac{1}{\alpha} \hat e_{\mu,\nu}(s,a,s') -1\Big) \text{ and }
    \red{\widehat{\bar{w}}}_{\mu}(x) = \exp(r(s,s')+\mu(s,s')-1). \label{eq:closed_form_solution_wss1_sample}
\end{align}
which is analogous to \eqref{eq:closed_form_solution_wsa_wss1}. Plugging this result into \eqref{eq:maximin_lagrangian_detail_sample} yields a sample-based objective function for minimization\footnote{
In contrast to the results in the previous sections where \eqref{eq:lagrange_objective_double} and \eqref{eq:maximin_lagrangian_detail} are identical, 
$\widehat{\gL}(\mu, \nu)$ in \eqref{eq:approx_lagrangian_objective_double} is an upper bound of $\max_{\blue{w}, \redb{w}} \widehat{\gL}(\blue{w},\redb{w},\mu,\nu)$ in \eqref{eq:maximin_lagrangian_detail_sample}, due to applying \emph{per-sample} closed-form solutions.
While unbiasedness has been sacrificed at the cost of eliminating the nested optimization, it enables much stable optimization in practice.
The upper bound gap vanishes when the transition dynamics are deterministic.}:
\begin{align}
    \min_{\mu,\nu}\, & \widehat \gL (\mu, \nu) 
    = (1-\gamma) \bE_{s_0 \in D_0}[\nu(s_0)] + \bE_{x \in D^I} \Big[\exp\big(r(s,s')+\mu(s,s')-1\big) \label{eq:approx_lagrangian_objective_double} \\
    & \hspace{20pt} + \alpha \exp\big(\tfrac{1}{\alpha} \hat e_{\mu,\nu}(s,a,s') - 1 \big) \Big]. \nonumber
\end{align}
Still, the variable $\mu \in \R^{S \times S}$ is much higher dimensional than $\nu \in \R^{S}$. So, it works as the main bottleneck for the overall optimization.
Fortunately, we can further simplify \eqref{eq:approx_lagrangian_objective_double} by eliminating its dependence on $\mu$ via exploiting an additional closed-form solution.
\begin{restatable}{proposition}{secondclosedformsolution}
For any $\nu$, the closed-form solution to the minimization \eqref{eq:approx_lagrangian_objective_double} with respect to $\mu$, i.e. $\mu_\nu = \argmin_{\mu} \widehat \gL(\mu, \nu)$, is
\begin{align}
    \mu_\nu(s,s') = \tfrac{1}{1+\alpha} \big(-\alpha r(s,s')+\gamma\nu(s')-\nu(s) \big).
    \label{eq:closed_form_solution_mu}
\end{align}
\end{restatable}
Using the above result in \eqref{eq:approx_lagrangian_objective_double}, we obtain the following minimization problem:
\begin{align}
    &\min_{\widehat\nu} \widehat\gL (\widehat\nu)
    =(1-\gamma) \bE_{s_0 \in D_0}[\widehat\nu(s_0)]  + (1+\alpha)\bE_{x \in D^I} \Big[ \exp \Big(\tfrac{1}{1+\alpha} \widehat A_{\widehat\nu}(s,a,s') -1 \Big) \Big], \label{eq:lagrange_objective_single}
\end{align}
where $\widehat A_\nu(s,a,s') := r(s,s')+\gamma\nu(s')-\nu(s)$.
The remaining issue is that optimizing \eqref{eq:lagrange_objective_single} is not practical because $\exp(\cdot)$
often causes numerical instability and gradient explosion.
To address this, we use a numerically-stable alternative of \eqref{eq:lagrange_objective_single}:

\begin{restatable}{proposition}{fencheldual}
Let $\widetilde \gL(\widetilde\nu)$ be the function:
\begin{align}
    &\widetilde \gL (\widetilde\nu) = (1-\gamma) \bE_{s_0 \in D_0}[\widetilde\nu(s_0)] +(1+\alpha) \log \bE_{x \in D^I} \Big[\exp\Big(\tfrac{1}{1+\alpha} \widehat A_{\widetilde\nu}(s,a,s')\Big)\Big].
    \label{eq:main_fenchel_objective}
\end{align}
Then, $\min_{\widehat \nu} \widehat \gL(\widehat \nu) = \min_{\widetilde \nu} \widetilde \gL(\widetilde \nu)$ holds.
Also, $\widetilde \gL(\widetilde\nu)$ is convex with respect to $\widetilde\nu$.
\label{prop:fencheldual}
\end{restatable}
In order see why minimizing $\widetilde \gL(\widetilde\nu)$ no longer suffers from numerical instability, 
note that the gradient 
$\nabla_x\log\E_{x\sim p}[\exp(h(x))]=\E_{x\sim p}[\frac{\exp(h(x))}{\E_{\bar{x}\sim p}[\exp(h(\bar{x}))]}\nabla_x h(x)]$
normalizes $\exp(\cdot)$ by softmax and thus tames large numerical values.
Finally, we can show the following connection between $\widehat\nu^*$ and $\widetilde\nu^*$:
\begin{restatable}{proposition}{ldfdconnection}
Let $\widehat{V}$ and $\widetilde{V}$ be the sets $\argmin_{\widehat\nu} \widehat \gL (\widehat\nu)$ and $\argmin_{\widetilde\nu} \widetilde \gL(\widetilde\nu)$, respectively. Then,
$\widetilde{V}=\{\widehat\nu^*+C\mid \widehat\nu^*\in \widehat{V}, C\in\R\}$
holds.
\label{prop:ld_fd_connection}
\end{restatable}
Proposition~\ref{prop:ld_fd_connection} implies that an unnormalized stationary distribution corrections of an optimal policy would be obtained from $\widetilde\nu^* \in \widetilde{V}$.
\begin{align*}
    \blue{\widehat w}_{\mu_{\widehat\nu^*}, {\widehat\nu^*}}(x)
    &= \exp\Big(\tfrac{1}{\alpha} \big( -\mu_{\widehat\nu^*}(s,s') + \gamma \widehat\nu^*(s') - \widehat\nu^*(s) \big) - 1 \Big) \\
    &= \exp\Big(\tfrac{1}{1 + \alpha} \widehat A_{\textcolor{purple}{\widehat\nu^*}}(s,a,s') - 1 \Big) & \text{(by \eqref{eq:closed_form_solution_mu})} \\
    &\propto \exp\Big(\tfrac{1}{1+\alpha} \widehat A_{\textcolor{purple}{\widetilde \nu^*}}(s,a,s') \Big)
    =: \widetilde w_{\widetilde \nu^*}(x) &\text{(by Proposition~\ref{prop:ld_fd_connection})}
\end{align*}

\paragraph{Policy extraction}
We must take caution when using $\widetilde w_{\widetilde \nu^*}(x)$ since it is an unnormalized density ratio, i.e. $\E_{x \sim d^I}[ \widetilde w_{\widetilde \nu^*}(x) ] \neq 1$.
Therefore, to extract a policy, we perform weighted BC using self-normalized importance sampling~\citep{owen2013snis}:
\begin{align}
    &\max_{\pi} 
    \tfrac
    {\sum_{x \in D^I} \widetilde w_{\widetilde \nu^*}(x) \log \pi(a|s)}
    {\sum_{x \in D^I} \widetilde w_{\widetilde \nu^*}(x)},
    \label{eq:self_normalized_weighted_bc}
\end{align}
which completes the derivation of the practical version of LobsDICE.
To sum up, LobsDICE solves $\widetilde \nu^* = \argmin_{\widetilde\nu} \widetilde \gL(\widetilde\nu)$ of \eqref{eq:main_fenchel_objective} via gradient descent, and extracts a policy via self-normalized weighted BC of \eqref{eq:self_normalized_weighted_bc}. Pseudocode for LobsDICE can be found in Appendix~\ref{app:pseudocode_continuous}.


\section{Related Work}
\paragraph{Learning from observation (LfO)}
Recent approaches for LfO are mostly \textit{on-policy}~\citep{liu2018ifovideo,torabi2018gaifo,yang2019iddm,sun2019fail,liu2019sail} algorithms and are not directly applicable to the offline LfO setting considered in this work.
MobILE~\citep{kidambi2021mobile} is a model-based LfO algorithm, but it encourages uncertainty for online exploration, which is not suitable for the offline setting.
BCO~\citep{torabi2018bco} uses an inverse dynamics model (IDM) to infer the missing expert actions and performs BC on the generated expert's state-action dataset.
In addition to common issues by vanilla BC, BCO is not guaranteed to recover the expert's behavior in general.
OPOLO~\citep{ZhuEtal2020opolo} is a principled off-policy LfO algorithm, but
it solves a nested optimization and requires evaluation on OOD action values during training, which suffers from numerical instability in the offline setting.
IQ-Learn~\citep{garg2021iq} solves online and offline IL problems while avoiding adversarial training by learning a single Q-function.
However, it also suffers from numerical instability in the offline setting by overestimating $Q$ due to using OOD action values.
RCE~\citep{eysenbach2021replacing} aims to solve example-based control tasks in which a collection of example success states is assumed to be provided instead of entire expert trajectories. While RCE can be applied to LfO in principle, it tends to stay in a few expert states that are easy to reach in the offline setting, as discussed in Section C.2 of~\citep{eysenbach2021replacing}.

\paragraph{Stationary DIstribution Correction Estimation (DICE)}
DICE-family algorithms perform stationary distribution estimation, and many of them have been proposed for off-policy evaluation~\citep{nachum2019dualdice,zhang2019gendice,zhang2020gradientdice,yang2020offpolicy,dai2020coindice}.
Other lines of works consider reinforcement learning~\citep{nachumE2019algaedice,lee2021optidice,lee2022coptidice}, offline policy selection~\citep{yang2020offline}.
ValueDICE~\citep{kostrikov2020valuedice} and OPOLO~\citep{ZhuEtal2020opolo} derive off-policy IL and LfO objectives using DICE. However, they suffer from numerical instability in the offline setting due to the nested min-max optimization and OOD action evaluation.
DemoDICE~\citep{kim2022demodice} is an offline IL algorithm that directly optimizes stationary distribution as ours and reduces to solving a convex minimization. Yet, it requires expert action labels and is not directly applicable to LfO.
Concurrent to our work, SMODICE~\citep{ma2022smodice} is an offline LfO method, aiming to match stationary \emph{state} distributions by minimizing an objective similar to OPOLO~\cite{ZhuEtal2020opolo} in \eqref{eq:opolo_upper_bound}. 
Thus it also relies on potentially loose upper bound of the divergence. 
In addition, matching the state distributions may not be sufficient to recover the expert's behavior, on which we provide detailed discussion in Appendix~\ref{app:why_ss_matching} and Remark~\ref{remark:smodice}.

\section{Experiments}
\label{sec:experiments}
In this section, we evaluate LobsDICE on both tabular and continuous MDPs.
We use four baseline methods in tabular MDPs:
BC on imperfect demonstrations, BCO~\citep{torabi2018bco}, and OPOLO~\citep{ZhuEtal2020opolo}.
Additionally, we designed a strong baseline DemoDICEfO, which extends the state-of-the-art offline IL algorithm, DemoDICE~\citep{kim2022demodice}. 
DemoDICEfO trains an inverse dynamics model, uses it to fill the missing actions in the expert demonstrations, and then runs DemoDICE using the approximate expert demonstrations and the imperfect demonstrations.
For continuous control tasks, we use two additional baselines: IQ-Learn~\citep{garg2021iq} and RCE~\citep{eysenbach2021replacing}.

\subsection{Random MDPs}
\label{subsec:random_mdps}
\begin{figure*}[t]
\centering
\includegraphics[width=\linewidth]{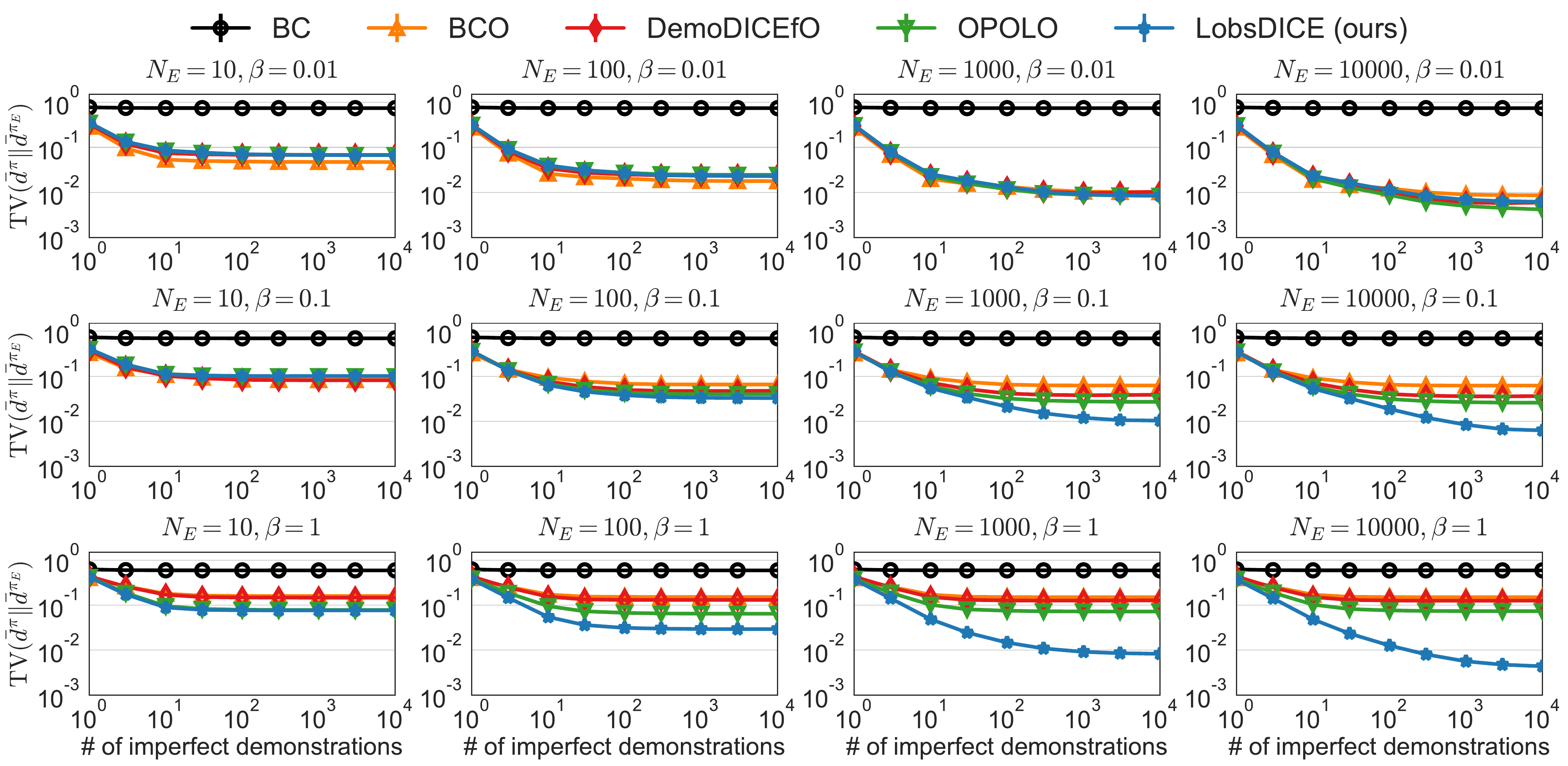}
\caption{
Performance of tabular LobsDICE and baselines in randomly generated MDPs. 
The first row indicates near-deterministic dynamics and the last row indicates highly stochastic dynamics.
As the level of stochasticity increases, baselines fall into suboptimal, even the number of state-only expert demonstrations and imperfect demonstrations increases, while LobsDICE goes to optimal.
For each algorithm, we measure the performance using total variation between state-transition stationary distributions of expert and learned policy.
We plot the mean and standard error of total variations $\mathrm{TV}(\bar{d}^\pi(s,s'),\bar{d}^{\pi_E}(s,s'))$ over 1000 random seeds.}
\label{fig:random_mdp}
\end{figure*}

We first evaluate LobsDICE and baseline algorithms on randomly generated finite MDPs using a varying number of expert/imperfect trajectories and different degrees of stochasticity in the environment .
We follow the experimental protocol in previous offline RL works~\citep{laroche2019spibb,lee2020bopah,lee2021optidice} but with additional control on the stochasticity of transition probabilities.
We conduct repeated experiments for 1K runs. 
For each run, (1) a random MDP is generated, (2) expert trajectories and imperfect trajectories are collected, and (3) each offline LfO algorithm is tested on the collected offline dataset.
We evaluate the performance of each algorithm by measuring the total variation distance between state-transition stationary distributions by the expert policy and the learned policy, i.e. $D_\mathrm{TV}(\bar d^\pi(s,s') \| \bar d^{\pi_E}(s,s'))$.
For the tabular MDP experiments, we adopt tabular methods but not function approximation.
Our tabular LobsDICE optimizes \eqref{eq:lagrange_objective_double} on the empirical MDP model constructed from the action-labeled dataset $D^I$ while extracting the policy through~\eqref{eq:policy_extraction_wbc}. 
The pseudocode for tabular LobsDICE can be found in Appendix~\ref{app:pseudocode_tabular}.
The detailed experimental setup such as random MDP generation and offline dataset generation can be found in Appendix~\ref{app:experiment_detail_random_mdp}.

\paragraph{Results}
Figure~\ref{fig:random_mdp} presents the results in random MDP experiments, where $\beta$ is the hyperparameter that controls the stochasticity of the underlying MDP.
The first row corresponds to the case when $\beta=0.01$ (nearly deterministic MDP).
In this situation, the inverse dynamics disagreement will be close to zero, i.e. $\KL(d^{\pi_1}(a |s,s') || d^{\pi_2}(a|s,s')) \approx 0$ for any two policies $\pi_1$ and $\pi_2$.
Thus, the algorithms whose performance directly relies on the IDM's accuracy (i.e. BCO and DemoDICEfO) even perform very well since it is very easy to learn a perfect inverse dynamics model in this scenario.
OPOLO's upper bound gap \eqref{eq:opolo_gap} will also be close to zero, thus OPOLO directly minimizes the divergence of state-transition distributions.
As a result, there is no performance gap among different algorithms, except for BC whose performance is determined by the quality of imperfect demonstrations.
Also, the performance of all algorithms (except for BC) improves as more data is given, which is natural.

The second row and the third row in Figure~\ref{fig:random_mdp} presents the result when $\beta=0.1$ (weakly stochastic MDP) and $\beta = 1.0$ (highly stochastic MDP) respectively.
In the stochastic MDPs, the IDM trained by the \emph{imperfect} demonstrations faces a challenge in predicting the expert's actions accurately (more challenging as $\beta$ gets larger). 
As a consequence, BCO gets suboptimal and its suboptimality cannot be improved even if more data is given.
DemoDICEfO performs better than BCO since it additionally considers the distributional shift by considering state distribution matching.
However, it is still suboptimal due to its nature that directly depends on the quality of inferred action by the learned IDM.
OPOLO does not rely on the learned IDM and outperforms both BCO and DemoDICEfO. 
Still, OPOLO can be inherently suboptimal due to its nature of optimizing the \emph{upper bound} unless the underlying transition dynamics are deterministic and injective.
This upper bound gap \eqref{eq:opolo_gap} is not controllable by the algorithm and implies that OPOLO can be suboptimal even given an infinite amount of data with sufficient dataset coverage, which can be seen in the rightmost figures.
Finally, our tabular LobsDICE using \eqref{eq:lagrange_objective_double} essentially solves the exact state-transition distribution matching problem (as $\alpha \rightarrow 0$). 
LobsDICE is the only offline LfO algorithm that can asymptotically recovers the expert's state demonstrations even though the underlying MDP is stochastic.

\subsection{Continuous control tasks (Gym-MuJoCo)}
\label{subsec:experiments_mujoco}
\begin{figure*}[t]
\centering
\includegraphics[width=\linewidth]{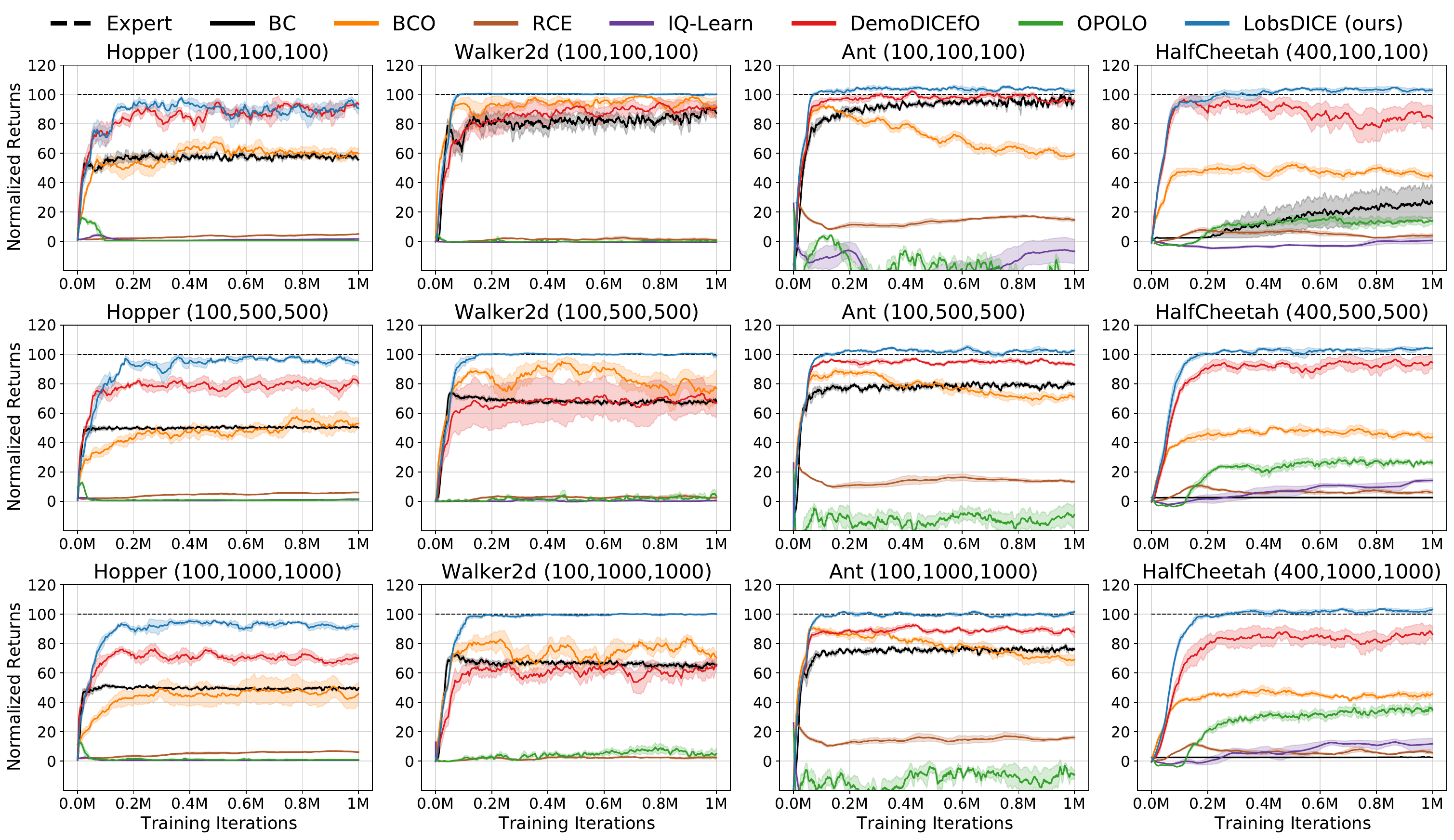}

\caption{
Performance of LobsDICE and baseline algorithms on various MuJoCo control tasks.
We build state-only expert demonstrations using 5 trajectories from \texttt{expert-v2}. 
For each task $(X,Y,Z)$ we construct imperfect demonstrations using
$X$, $Y$, and $Z$ trajectories from \texttt{expert-v2}, \texttt{medium-v2}, and \texttt{random-v2}, respectively.
We plot the mean and the standard errors (shaded area) of the normalized scores over five random seeds.}
\label{fig:mujoco}
\end{figure*}

We present the empirical performance of LobsDICE and baselines on MuJoCo~\citep{todorov2012mujoco} continuous control tasks using the OpenAI Gym~\citep{brockman2016gym} framework.
We utilize the D4RL dataset~\cite{fu2020d4rl} for offline LfO tasks in four MuJoCo environments: Hopper, Walker2d, HalfCheetah, and Ant.
Implementation details for LobsDICE and baseline algorithms such as hyperparameters and evaluation metric are provided in Appendix~\ref{app:experiment_detail_mujoco}.

\paragraph{Task setup}
For each MuJoCo environments, we employ \texttt{expert-v2}, \texttt{medium-v2}, and \texttt{random-v2} from D4RL datasets~\citep{fu2020d4rl}.
Across all environments, we consider three tasks, each of which uses different imperfect demonstrations while sharing the same expert observations.
First, we construct the state-only expert demonstration $D^E=\{(s,s')_i\}_{i=1}^{N_E}$ using the first 5 trajectories in \texttt{expert-v2}.
Then, we use trajectories in \texttt{expert-v2}, \texttt{medium-v2}, and \texttt{random-v2} to construct imperfect demonstrations with different ratios.
We denote the composition of imperfect demonstrations as $(X,Y,Z)$ in the title of each subplot in Figure~\ref{fig:mujoco}, which means that the imperfect dataset consists of $X$ trajectories from \texttt{expert-v2}, $Y$ trajectories from \texttt{medium-v2}, and $Z$ trajectories from \texttt{random-v2}.

\paragraph{Results}
Figure~\ref{fig:mujoco} summarizes that the empirical results of LobsDICE and baselines on continuous control tasks.
We first remark that LobsDICE (blue) significantly outperforms OPOLO (green) in all tasks across all domains, although both LobsDICE and OPOLO are DICE-based algorithms.
The failure of OPOLO comes from its numerical instability due to its dependence on nested optimization and using OOD action values during training. 
In contrast, LobsDICE solves a single minimization \eqref{eq:main_fenchel_objective} while it does not involve any evaluation on OOD actions, thus it is optimized stably.
IQ-Learn (purple) also suffers from numerical instability due to the usage of OOD action values during training, showing poor performance similar to OPOLO.
RCE (brown) tends to stay in a few states that are easy to reach in the offline setting, rather than following the entire expert trajectories.
Naive BC on imperfect demonstrations (black) is inherently suboptimal since it does not consider distribution-matching with the expert's observation at all.
While BCO (orange) exploits the expert's demonstrations with the inferred actions by the IDM, its policy learning is done only on the very scarce expert dataset (i.e. 5 trajectories), which makes the algorithm perform not well.
DemoDICEfO (red) exploits both expert demonstrations (where the missing actions are filled with the IDM) and the abundant imperfect demonstrations, but its performance is affected by the quality of the learned IDM. 
We empirically observe that the IDM error (on the true expert data) increases as the proportion of the non-expert data (i.e. \texttt{medium-v2} and \texttt{random-v2}) increases, resulting in performance degradation of both BCO and DemoDICEfO.
Finally, LobsDICE is the only algorithm that was able to fully recover the expert's performance regardless of the increase of non-expert data in the imperfect demonstrations, significantly outperforming baseline algorithms.
This result highlights the effectiveness of our method that solves a state-transition stationary matching problem in a principled manner.
We provide additional experiments in Appendix~\ref{app:additinoal_experiments}.

\section{Conclusion}
\label{sec:conc}
We presented LobsDICE, an algorithm for offline learning from observations (LfO), which successfully achieves state-of-the-art performance on various tabular and continuous tasks.
We formulated the offline LfO as a state-transition stationary distribution matching problem, where the stationary distribution is optimized via 
convex minimization.
Experimental results demonstrated that LobsDICE achieves promising performance in both tabular and continuous offline LfO tasks.

\begin{ack}
This work was supported by National Research Foundation (NRF) of Korea (NRF-2019R1A2C1087634),
Field-oriented Technology Development Project for Customs Administration through National Research Foundation (NRF) of Korea funded by the Ministry of Science \& ICT and Korea Customs Service (NRF-2021M3I1A1097938),
Institute of Information \& communications Technology Planning \& Evaluation (IITP) grant funded by the Korea government (MSIT) (No.2020-0-00940, Foundations of Safe Reinforcement Learning and Its Applications to Natural Language Processing; No.2022-0-00311, Development of Goal-Oriented Reinforcement Learning Techniques for Contact-Rich Robotic Manipulation of Everyday Objects; No.2019-0-00075, Artificial Intelligence Graduate School Program (KAIST); No.2021-0-02068, Artificial Intelligence Innovation Hub),
and Electronics and Telecommunications Research Institute (ETRI) grant funded by the Korean government (22ZS1100, Core Technology Research for Self-Improving Integrated Artificial Intelligence System).
Hongseok Yang was supported by the Engineering Research Center Program through the National Research Foundation of Korea (NRF) funded by the Korean Government MSIT (NRF-2018R1A5A1059921) and also by the Institute for Basic Science (IBS-R029-C1).
Kee-Eung Kim was supported by KAIST-NAVER Hypercreative AI Center.
\end{ack}

\bibliographystyle{abbrvnat}
\bibliography{neurips_2022}

\begin{thebibliography}{45}
\providecommand{\natexlab}[1]{#1}
\providecommand{\url}[1]{\texttt{#1}}
\expandafter\ifx\csname urlstyle\endcsname\relax
  \providecommand{\doi}[1]{doi: #1}\else
  \providecommand{\doi}{doi: \begingroup \urlstyle{rm}\Url}\fi

\bibitem[Argall et~al.(2009)Argall, Chernova, Veloso, and
  Browning]{argall2009survey}
B.~D. Argall, S.~Chernova, M.~Veloso, and B.~Browning.
\newblock A survey of robot learning from demonstration.
\newblock \emph{Robotics and Autonomous Systems}, 57\penalty0 (5):\penalty0
  469--483, 2009.

\bibitem[Bentivegna et~al.(2004)Bentivegna, Atkeson, and
  Cheng]{Bentivegna2004robot}
D.~C. Bentivegna, C.~G. Atkeson, and G.~Cheng.
\newblock Learning tasks from observation and practice.
\newblock \emph{Robotics and Autonomous Systems}, 47\penalty0 (2):\penalty0
  163--169, 2004.
\newblock Robot Learning from Demonstration.

\bibitem[Brockman et~al.(2016)Brockman, Cheung, Pettersson, Schneider,
  Schulman, Tang, and Zaremba]{brockman2016gym}
G.~Brockman, V.~Cheung, L.~Pettersson, J.~Schneider, J.~Schulman, J.~Tang, and
  W.~Zaremba.
\newblock Openai gym.
\newblock \emph{arXiv preprint arXiv:1606.01540}, 2016.

\bibitem[Dai et~al.(2020)Dai, Nachum, Chow, Li, Szepesvari, and
  Schuurmans]{dai2020coindice}
B.~Dai, O.~Nachum, Y.~Chow, L.~Li, C.~Szepesvari, and D.~Schuurmans.
\newblock {CoinDICE}: Off-policy confidence interval estimation.
\newblock In \emph{Advances in Neural Information Processing Systems
  (NeurIPS)}, volume~33, pages 9398--9411, 2020.

\bibitem[Eysenbach et~al.(2021)Eysenbach, Levine, and
  Salakhutdinov]{eysenbach2021replacing}
B.~Eysenbach, S.~Levine, and R.~R. Salakhutdinov.
\newblock Replacing rewards with examples: Example-based policy search via
  recursive classification.
\newblock \emph{Advances in Neural Information Processing Systems (NeurIPS)},
  34, 2021.

\bibitem[Fu et~al.(2020)Fu, Kumar, Nachum, Tucker, and Levine]{fu2020d4rl}
J.~Fu, A.~Kumar, O.~Nachum, G.~Tucker, and S.~Levine.
\newblock D4{RL}: Datasets for deep data-driven reinforcement learning, 2020.

\bibitem[Fujimoto et~al.(2019)Fujimoto, Meger, and Precup]{fujimoto2019bcq}
S.~Fujimoto, D.~Meger, and D.~Precup.
\newblock Off-policy deep reinforcement learning without exploration.
\newblock In \emph{International Conference on Machine Learning (ICML)}, pages
  2052--2062. PMLR, 2019.

\bibitem[Garg et~al.(2021)Garg, Chakraborty, Cundy, Song, and
  Ermon]{garg2021iq}
D.~Garg, S.~Chakraborty, C.~Cundy, J.~Song, and S.~Ermon.
\newblock Iq-learn: Inverse soft-q learning for imitation.
\newblock \emph{Advances in Neural Information Processing Systems (NeurIPS)},
  34, 2021.

\bibitem[Goodfellow et~al.(2014)Goodfellow, Pouget-Abadie, Mirza, Xu,
  Warde-Farley, Ozair, Courville, and Bengio]{goodfellow2014generative}
I.~J. Goodfellow, J.~Pouget-Abadie, M.~Mirza, B.~Xu, D.~Warde-Farley, S.~Ozair,
  A.~C. Courville, and Y.~Bengio.
\newblock Generative adversarial nets.
\newblock In \emph{Advances in Neural Information Processing Systems
  (NeurIPS)}, 2014.

\bibitem[Gulrajani et~al.(2017)Gulrajani, Ahmed, Arjovsky, Dumoulin, and
  Courville]{gulrajani2017improvedgan}
I.~Gulrajani, F.~Ahmed, M.~Arjovsky, V.~Dumoulin, and A.~Courville.
\newblock Improved training of {W}asserstein {GAN}s.
\newblock \emph{arXiv preprint arXiv:1704.00028}, 2017.

\bibitem[Ho and Ermon(2016)]{ho2016gail}
J.~Ho and S.~Ermon.
\newblock Generative adversarial imitation learning.
\newblock In \emph{Advances in Neural Information Processing Systems
  (NeurIPS)}, 2016.

\bibitem[Jin et~al.(2021)Jin, Yang, and Wang]{jin2021pessimism}
Y.~Jin, Z.~Yang, and Z.~Wang.
\newblock Is pessimism provably efficient for offline rl?
\newblock In \emph{Proceedings of the 38th International Conference on Machine
  Learning}, volume 139 of \emph{Proceedings of Machine Learning Research},
  pages 5084--5096, 18--24 Jul 2021.

\bibitem[Ke et~al.(2020)Ke, Choudhury, Barnes, Sun, Lee, and
  Srinivasa]{ke2020generalil}
L.~Ke, S.~Choudhury, M.~Barnes, W.~Sun, G.~Lee, and S.~Srinivasa.
\newblock Imitation learning as f-divergence minimization.
\newblock In \emph{International Workshop on the Algorithmic Foundations of
  Robotics}, pages 313--329. Springer, 2020.

\bibitem[Kidambi et~al.(2021)Kidambi, Chang, and Sun]{kidambi2021mobile}
R.~Kidambi, J.~D. Chang, and W.~Sun.
\newblock Mob{ILE}: Model-based imitation learning from observation alone.
\newblock In A.~Beygelzimer, Y.~Dauphin, P.~Liang, and J.~W. Vaughan, editors,
  \emph{Advances in Neural Information Processing Systems}, 2021.
\newblock URL \url{https://openreview.net/forum?id=_Rtm4rYnIIL}.

\bibitem[Kim et~al.(2022)Kim, Seo, Lee, Jeon, Hwang, Yang, and
  Kim]{kim2022demodice}
G.-H. Kim, S.~Seo, J.~Lee, W.~Jeon, H.~Hwang, H.~Yang, and K.-E. Kim.
\newblock Demo{DICE}: Offline imitation learning with supplementary imperfect
  demonstrations.
\newblock In \emph{International Conference on Learning Representations
  (ICLR)}, 2022.

\bibitem[Kostrikov et~al.(2019)Kostrikov, Agrawal, Dwibedi, Levine, and
  Tompson]{kostrikov2019discriminator}
I.~Kostrikov, K.~K. Agrawal, D.~Dwibedi, S.~Levine, and J.~Tompson.
\newblock Discriminator-actor-critic: Addressing sample inefficiency and reward
  bias in adversarial imitation learning.
\newblock In \emph{International Conference on Learning Representations
  (ICLR)}, 2019.

\bibitem[Kostrikov et~al.(2020)Kostrikov, Nachum, and
  Tompson]{kostrikov2020valuedice}
I.~Kostrikov, O.~Nachum, and J.~Tompson.
\newblock Imitation learning via off-policy distribution matching.
\newblock In \emph{International Conference on Learning Representations
  (ICLR)}, 2020.

\bibitem[Kumar et~al.(2019)Kumar, Fu, Soh, Tucker, and Levine]{kumar2019bear}
A.~Kumar, J.~Fu, M.~Soh, G.~Tucker, and S.~Levine.
\newblock Stabilizing off-policy {Q}-learning via bootstrapping error
  reduction.
\newblock In \emph{Advances in Neural Information Processing Systems
  (NeurIPS)}, 2019.

\bibitem[Kumar et~al.(2020)Kumar, Zhou, Tucker, and Levine]{kumar2020cql}
A.~Kumar, A.~Zhou, G.~Tucker, and S.~Levine.
\newblock Conservative {Q}-learning for offline reinforcement learning.
\newblock In \emph{Advances in Neural Information Processing Systems
  (NeurIPS)}, 2020.

\bibitem[Laroche et~al.(2019)Laroche, Trichelair, and
  Des~Combes]{laroche2019spibb}
R.~Laroche, P.~Trichelair, and R.~T. Des~Combes.
\newblock Safe policy improvement with baseline bootstrapping.
\newblock In \emph{International Conference on Machine Learning (ICML)}, pages
  3652--3661. PMLR, 2019.

\bibitem[Lee et~al.(2020)Lee, Lee, Vrancx, Kim, and Kim]{lee2020bopah}
B.~Lee, J.~Lee, P.~Vrancx, D.~Kim, and K.-E. Kim.
\newblock Batch reinforcement learning with hyperparameter gradients.
\newblock In \emph{International Conference on Machine Learning (ICML)}, pages
  5725--5735. PMLR, 2020.

\bibitem[Lee et~al.(2021)Lee, Jeon, Lee, Pineau, and Kim]{lee2021optidice}
J.~Lee, W.~Jeon, B.-J. Lee, J.~Pineau, and K.-E. Kim.
\newblock {OptiDICE}: Offline policy optimization via stationary distribution
  correction estimation.
\newblock In \emph{International Conference on Machine Learning (ICML)}, 2021.

\bibitem[Lee et~al.(2022)Lee, Paduraru, Mankowitz, Heess, Precup, Kim, and
  Guez]{lee2022coptidice}
J.~Lee, C.~Paduraru, D.~J. Mankowitz, N.~Heess, D.~Precup, K.-E. Kim, and
  A.~Guez.
\newblock {CO}pti{DICE}: Offline constrained reinforcement learning via
  stationary distribution correction estimation.
\newblock In \emph{International Conference on Learning Representations}, 2022.
\newblock URL \url{https://openreview.net/forum?id=FLA55mBee6Q}.

\bibitem[Li et~al.(2022)Li, Xu, Yu, and Luo]{li2022rethinking}
Z.~Li, T.~Xu, Y.~Yu, and Z.-Q. Luo.
\newblock Rethinking valuedice: Does it really improve performance?
\newblock \emph{arXiv preprint arXiv:2202.02468}, 2022.

\bibitem[Liu et~al.(2019)Liu, Ling, Mu, and Su]{liu2019sail}
F.~Liu, Z.~Ling, T.~Mu, and H.~Su.
\newblock State alignment-based imitation learning.
\newblock In \emph{International Conference on Learning Representations
  (ICLR)}, 2019.

\bibitem[Liu et~al.(2018)Liu, Gupta, Abbeel, and Levine]{liu2018ifovideo}
Y.~Liu, A.~Gupta, P.~Abbeel, and S.~Levine.
\newblock Imitation from observation: Learning to imitate behaviors from raw
  video via context translation.
\newblock In \emph{2018 IEEE International Conference on Robotics and
  Automation (ICRA)}, pages 1118--1125. IEEE, 2018.

\bibitem[Ma et~al.(2021)Ma, Jayaraman, and Bastani]{ma2021conservative}
Y.~Ma, D.~Jayaraman, and O.~Bastani.
\newblock Conservative offline distributional reinforcement learning.
\newblock \emph{Advances in Neural Information Processing Systems}, 34, 2021.

\bibitem[Ma et~al.(2022)Ma, Shen, Jayaraman, and Bastani]{ma2022smodice}
Y.~J. Ma, A.~Shen, D.~Jayaraman, and O.~Bastani.
\newblock Smodice: Versatile offline imitation learning via state occupancy
  matching.
\newblock \emph{arXiv preprint arXiv:2202.02433}, 2022.

\bibitem[Nachum et~al.(2019{\natexlab{a}})Nachum, Chow, Dai, and
  Li]{nachum2019dualdice}
O.~Nachum, Y.~Chow, B.~Dai, and L.~Li.
\newblock Dual{DICE}: Behavior-agnostic estimation of discounted stationary
  distribution corrections.
\newblock In \emph{Advances in Neural Information Processing Systems
  (NeurIPS)}, 2019{\natexlab{a}}.

\bibitem[Nachum et~al.(2019{\natexlab{b}})Nachum, Dai, Kostrikov, Chow, Li, and
  Schuurmans]{nachumE2019algaedice}
O.~Nachum, B.~Dai, I.~Kostrikov, Y.~Chow, L.~Li, and D.~Schuurmans.
\newblock Algae{DICE}: Policy gradient from arbitrary experience.
\newblock \emph{arXiv preprint arXiv:1912.02074}, 2019{\natexlab{b}}.

\bibitem[Owen(2013)]{owen2013snis}
A.~B. Owen.
\newblock \emph{Monte Carlo theory, methods and examples}.
\newblock 2013.

\bibitem[Puterman(1994)]{puterman1994markov}
M.~L. Puterman.
\newblock Markov decision processes: Discrete stochastic dynamic programming,
  1994.

\bibitem[Ross et~al.(2011)Ross, Gordon, and Bagnell]{ross2011reduction}
S.~Ross, G.~Gordon, and J.~A. Bagnell.
\newblock A reduction of imitation learning and structured prediction to
  no-regret online learning.
\newblock In \emph{Proceedings of the International Conference on Artificial
  Intelligence and Statistics (AISTATS)}, pages 627--635, 2011.

\bibitem[Schaal et~al.(1997)]{schaal1997lfd}
S.~Schaal et~al.
\newblock Learning from demonstration.
\newblock \emph{Advances in neural information processing systems (NeurIPS)},
  pages 1040--1046, 1997.

\bibitem[Sun et~al.(2019)Sun, Vemula, Boots, and Bagnell]{sun2019fail}
W.~Sun, A.~Vemula, B.~Boots, and D.~Bagnell.
\newblock Provably efficient imitation learning from observation alone.
\newblock In \emph{International conference on machine learning}, pages
  6036--6045. PMLR, 2019.

\bibitem[Sutton and Barto(1998)]{sutton1998rlbook}
R.~S. Sutton and A.~G. Barto.
\newblock \emph{Reinforcement learning: {A}n introduction}.
\newblock MIT Press, 1998.

\bibitem[Todorov et~al.(2012)Todorov, Erez, and Tassa]{todorov2012mujoco}
E.~Todorov, T.~Erez, and Y.~Tassa.
\newblock Mujoco: A physics engine for model-based control.
\newblock In \emph{2012 IEEE/RSJ International Conference on Intelligent Robots
  and Systems}, pages 5026--5033. IEEE, 2012.

\bibitem[Torabi et~al.(2018)Torabi, Warnell, and Stone]{torabi2018bco}
F.~Torabi, G.~Warnell, and P.~Stone.
\newblock Behavioral cloning from observation.
\newblock In \emph{International Joint Conference on Artificial Intelligence
  (IJCAI)}, 2018.

\bibitem[Torabi et~al.(2019)Torabi, Warnell, and Stone]{torabi2018gaifo}
F.~Torabi, G.~Warnell, and P.~Stone.
\newblock Generative adversarial imitation from observation.
\newblock \emph{ICML Workshop on Imitation, Intent, and Interaction (I3)},
  2019.

\bibitem[Yang et~al.(2019)Yang, Ma, Huang, Sun, Liu, Huang, and
  Gan]{yang2019iddm}
C.~Yang, X.~Ma, W.~Huang, F.~Sun, H.~Liu, J.~Huang, and C.~Gan.
\newblock Imitation learning from observations by minimizing inverse dynamics
  disagreement.
\newblock In \emph{Advances in Neural Information Processing Systems
  (NeurIPS)}, 2019.

\bibitem[Yang et~al.(2020{\natexlab{a}})Yang, Dai, Nachum, Tucker, and
  Schuurmans]{yang2020offline}
M.~Yang, B.~Dai, O.~Nachum, G.~Tucker, and D.~Schuurmans.
\newblock Offline policy selection under uncertainty, 2020{\natexlab{a}}.

\bibitem[Yang et~al.(2020{\natexlab{b}})Yang, Nachum, Dai, Li, and
  Schuurmans]{yang2020offpolicy}
M.~Yang, O.~Nachum, B.~Dai, L.~Li, and D.~Schuurmans.
\newblock Off-policy evaluation via the regularized lagrangian.
\newblock In \emph{Advances in Neural Information Processing Systems
  (NeurIPS)}, 2020{\natexlab{b}}.

\bibitem[Zhang et~al.(2020{\natexlab{a}})Zhang, Dai, Li, and
  Schuurmans]{zhang2019gendice}
R.~Zhang, B.~Dai, L.~Li, and D.~Schuurmans.
\newblock Gen{DICE}: Generalized offline estimation of stationary values.
\newblock In \emph{Proceedings of the 8th International Conference on Learning
  Representations (ICLR)}, 2020{\natexlab{a}}.

\bibitem[Zhang et~al.(2020{\natexlab{b}})Zhang, Liu, and
  Whiteson]{zhang2020gradientdice}
S.~Zhang, B.~Liu, and S.~Whiteson.
\newblock Gradient{DICE}: Rethinking generalized offline estimation of
  stationary values.
\newblock In \emph{Proceedings of the 35th International Conference on Machine
  Learning (ICML)}, 2020{\natexlab{b}}.

\bibitem[Zhu et~al.(2020)Zhu, Lin, Dai, and Zhou]{ZhuEtal2020opolo}
Z.~Zhu, K.~Lin, B.~Dai, and J.~Zhou.
\newblock Off-policy imitation learning from observations.
\newblock In \emph{Advances in Neural Information Processing Systems
  (NeurIPS)}, 2020.

\end{thebibliography}

\section*{Checklist}
\begin{enumerate}

\item For all authors...
\begin{enumerate}
  \item Do the main claims made in the abstract and introduction accurately reflect the paper's contributions and scope?
    \answerYes{} {See abstract and introduction (Section~\ref{sec:intro})}
  \item Did you describe the limitations of your work?
    \answerYes{} {Due to space limit, we discuss it in the supplementary material (Section~\ref{app:limitation})}
  \item Did you discuss any potential negative societal impacts of your work?
    \answerYes{} Due to space limit, we discuss it in the supplementary material (Section~\ref{app:social_impact}). 
  \item Have you read the ethics review guidelines and ensured that your paper conforms to them?
    \answerYes{}
\end{enumerate}

\item If you are including theoretical results...
\begin{enumerate}
  \item Did you state the full set of assumptions of all theoretical results?
    \answerYes{} In appropriate places.
        \item Did you include complete proofs of all theoretical results?
    \answerYes{} {Due to space limit, we provide proofs in the supplementary material. See Section~\ref{app:theoretical_analysis}}
\end{enumerate}
\item If you ran experiments...
\begin{enumerate}
  \item Did you include the code, data, and instructions needed to reproduce the main experimental results (either in the supplemental material or as a URL)?
    \answerYes{} We submit the code as supplementary material. 
  \item Did you specify all the training details (e.g., data splits, hyperparameters, how they were chosen)?
    \answerYes{} We specify them in the supplementary material. {See Section~\ref{app:experiment_detail}}
        \item Did you report error bars (e.g., with respect to the random seed after running experiments multiple times)?
    \answerYes{} {We report them in each figure. See Figure~\ref{fig:random_mdp} and~\ref{fig:mujoco}}. 
        \item Did you include the total amount of compute and the type of resources used (e.g., type of GPUs, internal cluster, or cloud provider)? 
    \answerYes{} {We specify them in the supplementary material. See Section~\ref{app:resource}}
\end{enumerate}

\item If you are using existing assets (e.g., code, data, models) or curating/releasing new assets...
\begin{enumerate}
  \item If your work uses existing assets, did you cite the creators?
    \answerYes{} {We cite Random MDP protocols and D4RL dataset. See Section~\ref{sec:experiments}}.
  \item Did you mention the license of the assets?
    \answerYes{} {D4RL dataset is licensed under the Apache 2.0. See Section~\ref{app:license}}.
  \item Did you include any new assets either in the supplemental material or as a URL?
    \answerNA{} 
  \item Did you discuss whether and how consent was obtained from people whose data you're using/curating?
    \answerNA{} All the datasets we used are public.
  \item Did you discuss whether the data you are using/curating contains personally identifiable information or offensive content?
    \answerNA{} None of datasets contain identifiable contents.
\end{enumerate}

\item If you used crowdsourcing or conducted research with human subjects...
\begin{enumerate}
  \item Did you include the full text of instructions given to participants and screenshots, if applicable?
    \answerNA{} Not applicable.
  \item Did you describe any potential participant risks, with links to Institutional Review Board (IRB) approvals, if applicable?
    \answerNA{} Not applicable.
  \item Did you include the estimated hourly wage paid to participants and the total amount spent on participant compensation?
    \answerNA{} Not applicable.
\end{enumerate}

\end{enumerate}

\appendix
\clearpage
\section{Why State-transition Occupancy Matching Instead of State Occupancy Matching?}
\label{app:why_ss_matching}
\begin{figure}[h]
    \centering
    \includegraphics[width=0.4\textwidth]{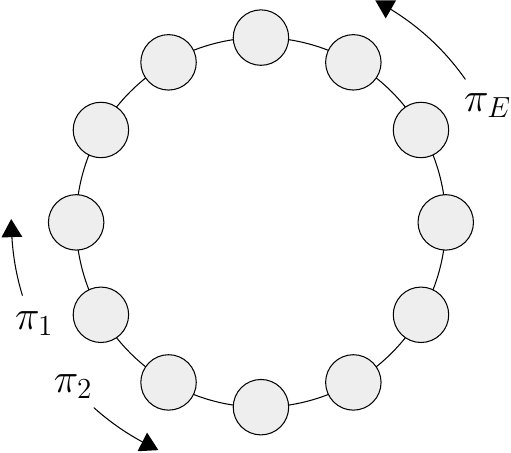}
    \caption{An example MDP with $|S|=12$, $|A|=2$ and $p_0(s) = \mathrm{Unif}(s)$. In this example MDP, $\{\red{\pi_1}, \pi_2\} \subseteq \argmin_{\pi} \KL(d^\pi(s) || d^{\pi_E}(s))$, while $\{\pi_2\} = \argmin_{\pi} \KL(d^\pi(s,s') || d^{\pi_E}(s,s'))$.}
    \label{fig:why_ss_matching}
\end{figure}

For learning from observation, we adopt the objective function for $d(s,s')$-matching (i.e., $\min_{\pi} \KL(d^\pi(s,s') || d^E(s,s'))$) instead of $d(s)$-matching (i.e. $\min_{\pi} \KL(d^\pi(s) || d^E(s))$) since $d(s)$-matching may ignore the \emph{directionality} of the expert trajectories.
To see this, consider an MDP with $|S|$ states and 2 actions, where states are denoted as circles in Figure~\ref{fig:why_ss_matching}.
Each action moves the agent from the current state to the neighboring state deterministically either clockwise or counterclockwise.
Initial state distribution $p_0$ is defined as uniform distribution over the entire states.
Finally, let an expert $\pi_E$ be a policy that moves in a counterclockwise direction, and we want to mimic this expert's behavior.
In this example, the desired imitation policy should be $\pi_2$ in Figure~\ref{fig:why_ss_matching}.
However, if we perform $d(s)$-matching, even $\pi_1$ can be obtained as a resulting policy, which is the complete opposite of what we wanted to obtain.
This is due to the fact that $\pi_1$ and $\pi_2$ share the same state stationary distribution of a uniform distribution: $d^\pi(s) = \frac{1}{|S|} ~ \forall s$. Thus, both $\pi_1$ and $\pi_2$ must be an optimal solution of $\min_{\pi} \KL(d^\pi(s) || d^E(s))$.
In contrast, the $d(s,s')$-matching has a unique solution of $\pi_2$, exploiting the directionality information (i.e. moving counterclockwise) of the expert's trajectories.

\section{LobsDICE for State Occupancy Matching}
\label{app:lobsdice_for_state_matching}

While we have provided a counter-example where $d(s)$-matching can fail to recover the expert's behavior in Appendix~\ref{app:why_ss_matching}, we still can derive an offline LfO algorithm for $d(s)$-matching.
For the given policy $\pi$, its state stationary distribution is defined as:
\begin{align*}
    \bar d^\pi(s) &:= (1-\gamma)\textstyle\sum\limits_{t=0}^\infty\gamma^t\Pr(s_t=s).
\end{align*}

\begin{figure*}[h!]
\centering
\vspace{-0.1cm}
\includegraphics[width=\linewidth]{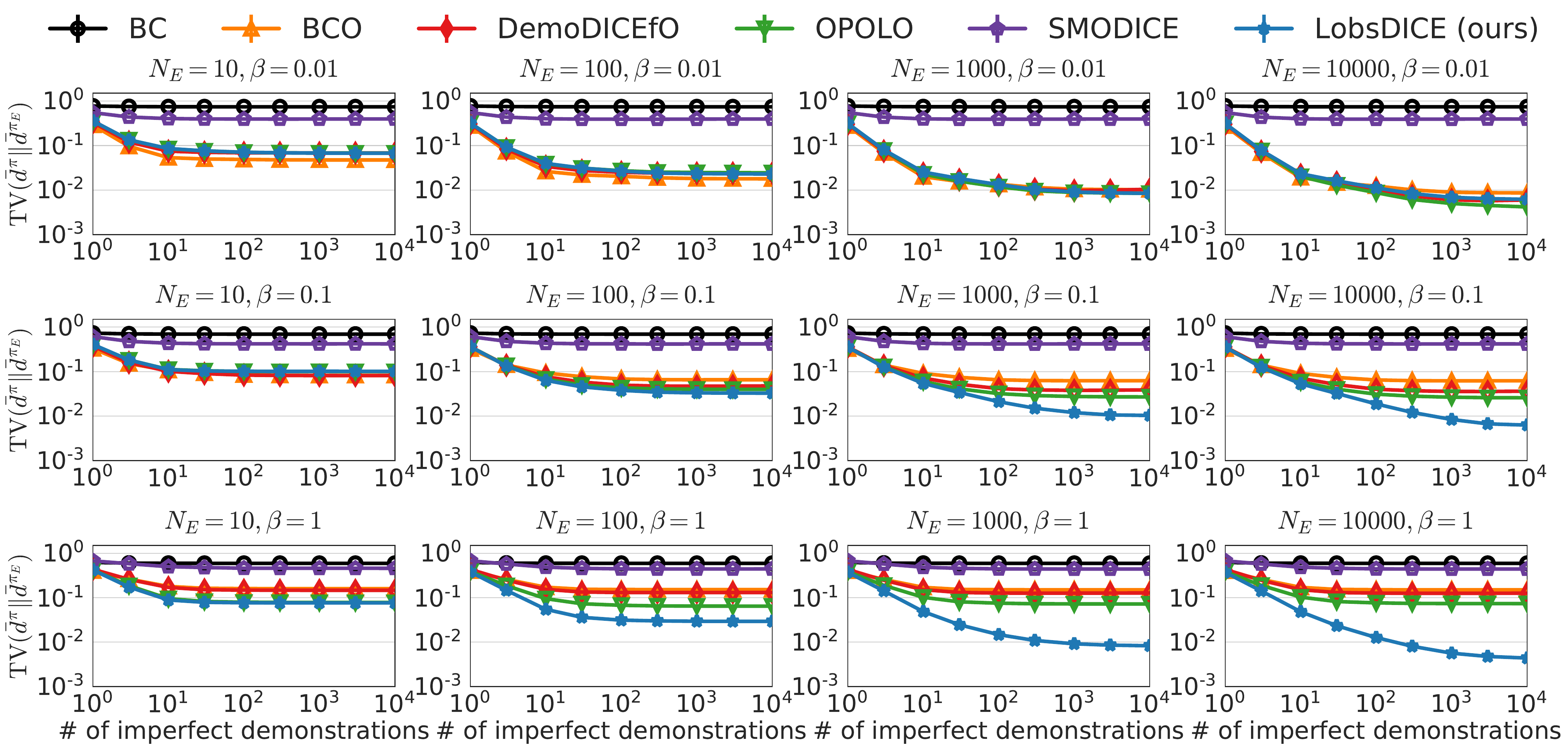}
\vspace{-0.6cm}
\caption{\revised{
Performance of tabular LobsDICE and baselines in randomly generated MDPs. 
The first row indicates near-deterministic dynamics and the last row indicates highly stochastic dynamics.
As the level of stochasticity increases, baselines fall into suboptimal, even the number of state-only expert demonstrations and imperfect demonstrations increases, while LobsDICE goes to optimal.
For each algorithm, we measure the performance using total variation between state-transition stationary distributions of expert and learned policy.
We plot the mean and standard error of total variations $\mathrm{TV}(\bar{d}^\pi(s,s'),\bar{d}^{\pi_E}(s,s'))$ over 1000 random seeds.}}
\label{fig:random_mdp_smodice}
\end{figure*}

Then, similar to the state-transition occupancy matching objective~\eqref{eq:state_transition_matching}, we formulate a state occupancy matching problem as follows:
\begin{equation}
    \min_\pi\KL(\red{\bar{d}^\pi(s)} \| {d}^E(s) ) + \alpha\KL( \blue{d^\pi(s,a)} \| d^I(s,a)).
    \label{eq:state_matching}
\end{equation}
where the hyperparameter $\alpha > 0$ balances between encouraging state matching and preventing distribution shift from the distribution of imperfect demonstrations.
This objective can be reformulated in terms of optimizing stationary distribution:
\begin{align}
    \max_{\blue{d},\redb{d}\ge0}~ 
    &-\KL(\redb{d}(s)\|{d}^E(s)) - \alpha\KL(\blue{d}(s,a) \| d^I(s,a)) \label{eq:state_main_objective}\\
    \text{s.t.}~
    &\textstyle\sum\limits_{a'} \blue{d}(s',a') = (1-\gamma)p_0(s') + \gamma \sum\limits_{s,a} \blue{d}(s,a) T(s'|s,a) \quad\forall s',\label{eq:state_bellman_flow_constraint}\\
    &\textstyle \sum\limits_{a}\blue{d}(s,a) = \redb{d}(s)\quad\forall s,\label{eq:state_marginalization_constraint}
\end{align}
Then, we consider the Lagrangian for the constrained optimization (\ref{eq:state_main_objective}-\ref{eq:state_marginalization_constraint}):
\begin{align}
    &\min_{\mu,\nu} \max_{\blue{d},\redb{d}\ge0} \textstyle -\E_{\redb{d}}\Big[\log\frac{\redb{d}(s)}{d^E(s)}\Big]-\E_{\blue{d}}\bigg[\log\frac{\blue{d}(s,a)}{d^I(s,a)}\bigg] 
    + \sum\limits_s \mu(s) \big( \redb{d}(s)-\sum\limits_a \blue{d}(s,a) \big) \label{eq:state_lagrangian} \\
    & \hspace{30pt}\textstyle+ \sum\limits_{s'} \nu(s') \big((1-\gamma)p_0(s') + \gamma\sum\limits_{s,a}\blue{d}(s,a)T(s'|s,a)-\sum_{a'}\blue{d}(s',a')\big) \nonumber
\end{align}
where $\nu(s') \in \R$ are the Lagrange multipliers for the Bellman flow constraints~\eqref{eq:state_bellman_flow_constraint}, and $\mu(s) \in \R$ are the Lagrange multipliers for the marginalization constraints~\eqref{eq:state_marginalization_constraint}.
To make the optimization tractable in the offline setting, we rearrange the terms in \eqref{eq:state_lagrangian}, introducing new optimization variables $w$ and $\bar w$ that denote stationary distribution correction ratios for $(s,a)$ and $s$ respectively.
\begin{align}
    \eqref{eq:state_lagrangian}=&\min_{\mu,\nu}\max_{\blue{d},\redb{d}\ge0} (1 - \gamma) \E_{s_0 \sim p_0} [\nu(s_0)]  
    + \E_{s\sim \redb{d}} \Big[ \mu(s) - \log\underbrace{\tfrac{\redb{d}(s)}{d^I(s)}}_{=:\redb{w}(s)} + \underbrace{\log\tfrac{d^E(s)}{d^I(s)}}_{=: r(s)} \Big] \nonumber \\
    & \hspace{30pt} + \E_{(s,a) \sim \blue{d}} \Big[ \underbrace{ - \mu(s) + \E_{s'}[\gamma \nu(s')] - \nu(s)}_{=:e_{\mu,\nu}(s,a)} - \alpha\log\underbrace{\tfrac{\blue{d}(s,a)}{d^I(s,a)}}_{=:\blue{w}(s,a)} \Big] \nonumber \\
    =& \min_{\mu,\nu} \max_{\blue{w},\redb{w}\ge0} (1 - \gamma) \E_{s_0 \sim p_0} [\nu(s_0)] + \E_{s\sim d^I} \big[ \redb{w}(s) \big( r(s) + \mu(s) - \log\redb{w}(s) \big)\big] \nonumber \\
    & \hspace{30pt} + \E_{(s,a) \sim d^I} \big[\blue{w}(s,a) \big( e_{\mu,\nu}(s,a) - \alpha\log\blue{w}(s,a) \big)\big] =: \gL(\blue{w},\redb{w},\mu,\nu).
    \label{eq:minimax_state_matching}
\end{align}
We introduce the log ratio $r(s)=\log\frac{d^E(s)}{d^I(s)}$ to make expectation for $d^I$ instead of $d^E$. The log ratio $r(s)$ can be estimated by using a pretrained discriminator $c:S\rightarrow[0,1]$, where $c$ is trained by:
\[
c^*=\argmax_{c:S\rightarrow[0,1]}\E_{s \sim d^E}[\log c(s)] + \E_{s \sim d^I}[\log c(s)].
\]
Then, it is proven that the optimal discriminator $c^*$ satisfies $c^*(s)=\frac{d^E(s)}{d^E(s)+d^I(s)}$.
Therefore, $r(s)$ can be estimated by:
\[
r(s)=-\log\bigg(\frac{1}{c^*(s)}-1\bigg).
\]
Similar to Proposition~\ref{prop:closed_form_solution}, we can easily derive the closed-form solution to the inner maximization in \eqref{eq:minimax_state_matching}:
\begin{align*}
    \blue{w}_{\mu,\nu}(s,a)&=\exp\Big(\tfrac{1}{\alpha}e_{\mu,\nu}(s,a)-1\Big),\\
    \redb{w}_\mu(s)&=\exp(r(s)+\mu(s)-1).
\end{align*}
Finally, by plugging these closed-form solutions into \eqref{eq:minimax_state_matching}, the nested min-max optimization of~\eqref{eq:minimax_state_matching} is reduced to a single minimization:
\begin{align}
    \min_{\mu,\nu}&\gL(\blue{w}_{\mu,\nu},\redb{w}_{\mu},\mu,\nu)=(1 - \gamma) \E_{s_0 \sim p_0} [\nu(s_0)] + \E_{s\sim d^I} \big[ \exp\big( r(s) + \mu(s) - 1 \big)\big] \label{eq:state_lagrange_objective_double} \\
    & \hspace{30pt} + \alpha\E_{(s,a) \sim d^I} \Big[ \exp\Big( \tfrac{1}{\alpha}e_{\mu,\nu}(s,a) - 1 \Big) \Big], \nonumber
\end{align}
which is a convex function of $\mu$ and $\nu$. Finally, once we obtain the optimal solution $(\mu^*, \nu^*)$ of \eqref{eq:state_lagrange_objective_double}, we have $\blue{w}_{\mu^*,\nu^*}(s,a) = \frac{d^{\pi^*}(s,a)}{d^I(s,a)}$. Then, we can extract an optimal policy via weighted BC:
\begin{align}
    \max_{\pi} \E_{(s,a) \sim d^{\pi^*}}[ \log \pi(a|s) ] = \E_{(s,a) \sim d^I} [ \blue{w}_{\mu^*,\nu^*}(s,a) \log \pi(a|s) ]
\end{align}
\begin{remark}
To the best of our knowledge, \eqref{eq:state_lagrange_objective_double} is the first result to directly solve the state distribution matching problem in an offline setting. Although SMODICE~\citep{ma2022smodice}, a concurrent work with ours, also aims to solve offline LfO in terms of optimizing stationary distribution, it essentially minimizes the (potentially loose) \emph{upper bound}, i.e., 
\begin{align}
    \min_{\pi} &\E_{s \sim d^\pi} \left[ \log \frac{d^I(s)}{d^E(s)} \right] + \KL(d^\pi(s,a) || d^I(s,a)) \label{eq:smodice_objective} \\
    \ge &\E_{s \sim d^\pi} \left[ \log \frac{d^I(s)}{d^E(s)} \right] + \KL(d^\pi(s) || d^I(s)) \\
    = &\KL(d^\pi(s) || d^E(s)) \label{eq:smodice_s_matching}
\end{align}
\revised{The upper bound gap $\eqref{eq:smodice_objective}-\eqref{eq:smodice_s_matching}$ is given by $\KL(\pi(a|s) || \pi^I(a|s))$. Unlike OPOLO of \eqref{eq:opolo_gap}, this upper bound gap does not vanish even when the transition is deterministic.}
Consequently, SMODICE may not be able to precisely recover the expert's state visitations even given an infinite amount of data due to the upper bound gap \revised{unless $d^I$ is collected purely by expert}.
In contrast, we are directly minimizing the divergence: solving \eqref{eq:state_lagrange_objective_double} is exactly equivalent to solving \eqref{eq:state_matching}.
\revised{As can be seen in Figure~\ref{fig:random_mdp_smodice}, SMODICE optimizing \eqref{eq:smodice_objective} fails to recover the expert's behavior due to the upper bound gap that is irreducible even for deterministic MDPs.}
\label{remark:smodice}
\end{remark}

\section{Validity of Weighted Behavior Cloning for Policy Extraction}
\label{app:policy_extraction_tabular}
For tabular MDPs, we can show that the weighted behavior cloning of \eqref{eq:policy_extraction_wbc}
\begin{align}
    &\max_{\pi} \E_{(s,a) \sim d^I} [ \blue{w}_{\mu^*, \nu^*}(s,a) \log \pi(a|s) ]  \tag{\ref{eq:policy_extraction_wbc}}
\end{align}
exactly yields an optimal policy $\pi^*$ of the original optimization problem \eqref{eq:state_transition_matching}.
First, the optimization problem \eqref{eq:state_transition_matching} and the constrained optimization problem (\ref{eq:main_objective}-\ref{eq:marginalization_constraint}) are equivalent with the following relationship between $\pi^*$ and $\blue{d}^*$:
\begin{align}
     \pi^*(a|s) = \frac{\blue{d}^*(s,a)}{\sum_{a'} \blue{d}^*(s,a')} 
     \label{eq:pi_d_relationship}
\end{align}
where $d^*$ is the optimal solution of both (\ref{eq:main_objective}-\ref{eq:marginalization_constraint}) and \eqref{eq:using_transpose_operator}.
Then, $\blue{w}_{\mu^*,\nu^*}(s,a)$ is computed by \eqref{eq:closed_form_solution_wsa_wss1} with $(\mu^*,\nu^*)$, the optimal solution of both \eqref{eq:lagrange_objective_double} and \eqref{eq:maximin_lagrangian_detail}.
Consequently, $\blue{w}_{\mu^*,\nu^*}(s,a) = \frac{\blue{d}^{*}(s,a)}{d^I(s,a)}$ holds, and thus the weighted behavior cloning of \eqref{eq:policy_extraction_wbc} essentially performs the following:
\begin{align}
    &\max_{x} \textstyle 
    \sum\limits_{s,a} \blue{d}^*(s,a) \log x(a|s) \label{eq:wbc_objective} \\
    &~\text{s.t. } \textstyle \sum\limits_{a} x(a|s) = 1 ~~~ \forall s \label{eq:wbc_constraint}
\end{align}
Now, consider the Lagrangian for the constrained optimization (\ref{eq:wbc_objective}-\ref{eq:wbc_constraint}):
\begin{align}
    \textstyle L := \sum\limits_{s,a} \blue{d}^*(s,a) \log x(a|s) - \sum\limits_{s} \lambda(s) \big( \sum\limits_{a} x(a|s) - 1 \big)
\end{align}
where $\lambda(s)$ is the Lagrange multiplier for the constraint \eqref{eq:wbc_constraint}. Then,
\begin{align}
    \frac{\partial L}{\partial x(a|s)} &= \frac{\blue{d}^*(s,a)}{x(a|s)} - \lambda(s) = 0  ~~~~ \Rightarrow x^*(a|s) = \frac{\blue{d}^*(s,a)}{\lambda(s)} ~ \forall s,a
\end{align}
Finally, considering the constraint $\sum_{a} x^*(a|s) = 1~ \forall s$, we obtain:
\begin{align}
    x^*(a|s) = \frac{\blue{d}^*(s,a)}{\sum_{a'} \blue{d}^*(s,a')}
\end{align}
which is identical to \eqref{eq:pi_d_relationship}.
In summary, the weighted behavior cloning performs the process of extracting the optimal policy $\pi^*$ from $w_{\mu^*,\nu^*}$.

\section{Challenges of Extending DemoDICE to Offline Learning from Observation}
\label{app:challenges_extending_demodice}
DemoDICE~\citep{kim2022demodice} is an algorithm that optimizes the state-action stationary distribution $d(s,a)$, where its naive application to $(s,s')$-distribution matching problem yields: 
\begin{align}
    \min_{d \ge 0} & \overbrace{\E_{\substack{(s,a) \sim d \\ s' \sim T(s,a)}} \Big[ \log \tfrac{\sum_{a} d(s,a) T(s'|s,a)}{d^E(s,s')} \Big]}^{= \KL( \bar d(s,s') || \bar d^E(s,s') )} + \alpha \overbrace{\E_{(s,a) \sim d} \Big[ \log \tfrac{d(s,a)}{d^I(s,a)} \Big]}^{= \KL(d(s,a) || d^I(s,a))} \\
    \text{s.t. } &\textstyle\sum\limits_{a'} d(s',a') = (1-\gamma)p_0(s') + \gamma \sum\limits_{s,a} d(s,a) T(s'|s,a) \quad\forall s',
\end{align}
However, estimating $\E_{\substack{(s,a) \sim d \\ s' \sim T(s,a)}} \Big[ \log \frac{\sum_{a} d(s,a) T(s'|s,a)}{d^E(s,s')} \Big]$ is intractable in the offline setting due to its inclusion of marginalization over $a$ inside the $\log (\cdot)$.
This challenge can be addressed by using the \emph{upper bound} proposed by OPOLO~\citep{ZhuEtal2020opolo}: 
\begin{align}
    \KL( \bar d^\pi(s,s') || d^E(s,s') \le \E_{(s,s') \sim \bar d^\pi(s,s')} \Big[ \log \tfrac{ \bar d^I(s,s') }{\bar d^E(s,s')} \Big] + \KL( d^\pi(s,a) \| d^I(s,a) ),
    \tag{\ref{eq:opolo_upper_bound}}
\end{align}
Replacing $\KL( \bar d^\pi(s,s') || d^E(s,s'))$ with the (potentially loose) \emph{upper bound} in the objective function yields the following tractable (but biased) optimization problem:
\begin{align}
    \min_{d \ge 0} & \E_{\substack{(s,a) \sim d \\ s' \sim T(s,a)}} \Big[ \log \tfrac{ \bar d^I(s,s') }{ \bar d^E(s,s')} \Big] + (1 + \alpha) \E_{(s,a) \sim d} \Big[ \log \tfrac{d(s,a)}{d^I(s,a)} \Big] \\
    \text{s.t. } &\textstyle\sum\limits_{a'} d(s',a') = (1-\gamma)p_0(s') + \gamma \sum\limits_{s,a} d(s,a) T(s'|s,a) \quad\forall s',
\end{align}

In contrast, we address the intractability challenge by introducing an additional optimization variable $\bar d(s,s')$ along with the marginalization constraint \eqref{eq:marginalization_constraint}. This allows for tractable offline optimization by (\ref{eq:maximin_lagrangian_detail},\ref{eq:lagrange_objective_double}) without introducing bias, which is our novel contribution.

\section{More Discussions on LobsDICE with Finite Samples}
\label{app:finite_lobsdice_derivation}

In this section, we provide additional discussions on LobsDICE with finite samples.

\subsection{LobsDICE with sample-based approximation: distribution matching on the MLE MDP}
We first show that LobsDICE that optimizes \eqref{eq:maximin_lagrangian_detail_sample} is equivalent to performing optimization of (\ref{eq:main_objective}-\ref{eq:marginalization_constraint}) on the MLE MDP constructed by $D^I$.
Let $\hat{M}\setminus R = \langle S, A, \hat{T}, \widehat p_0, \gamma \rangle$ be the MLE MDP constructed using $D^I$. Then, constrained optimization~(\ref{eq:main_objective}-\ref{eq:marginalization_constraint}) for the MLE MDP $\hat{M}\setminus R$ can be formulated as:
\begin{align}
    \max_{\blue{d},\redb{d}\ge0}~ 
    &-\KL(\redb{d}(s,s') \| \widehat{d}^E(s,s') ) - \alpha\KL(\blue{d}(s,a) \| \widehat{d}^I(s,a)) \label{eq:sample_based_obj}\\
    \text{s.t.}~
    &\textstyle\sum\limits_{a'} \blue{d}(s',a') = (1-\gamma)\widehat{p}_0(s') + \gamma \sum\limits_{s,a} \blue{d}(s,a) \hat{T}(s'|s,a) \quad\forall s',\label{eq:sample_based_const1}\\
    &\textstyle \sum\limits_{a}\blue{d}(s,a) \hat{T}(s'|s,a) = \redb{d}(s,s')\quad\forall s,s',\label{eq:sample_based_const2}
\end{align}
where $\widehat{p}_0$, $\widehat{d}^E$, and $\widehat{d}^I$ are empirical distributions of $D_0$, $D^E$, and $D^I$, respectively.
We consider Lagrangian for the above constrained optimization:
\begin{align*}
    &\min_{\mu,\nu} \max_{\blue{d},\redb{d}\ge0}
    - \E_{\redb{d}}\Big[ \log \tfrac{\redb{d}(s,s')}{\widehat{d}^E(s,s')} \Big] - \alpha \E_{\blue{d}}\Big[ \log \tfrac{\blue{d}(s,a)}{\widehat{d}^I(s,a)} \Big]
    + \textstyle\sum\limits_{s,s'} \mu(s,s') \big( \redb{d}(s,s') - \sum\limits_{a} \blue{d}(s,a) \hat{T}(s'|s,a) \big) \nonumber \\
    & + \textstyle\sum\limits_{s'} \nu(s') \big( (1-\gamma)\widehat{p}_0(s') + \gamma \sum\limits_{s,a} \blue{d} (s,a) \hat{T}(s'|s,a) - \sum\limits_{a'}\blue{d}(s',a') \big)\\
    =&\min_{\mu,\nu} \max_{\blue{d},\redb{d}\ge0} (1 - \gamma) \E_{s_0 \sim \widehat{p}_0} [\nu(s_0)]  
    + \E_{(s,s') \sim \redb{d}} \Big[ \mu(s,s') - \log\underbrace{\tfrac{\redb{d}(s,s')}{\widehat{d}^I(s,s')}}_{=:\redb{w}(s,s')} + \underbrace{\log\tfrac{\widehat{d}^E(s,s')}{\widehat{d}^I(s,s')}}_{=: \widehat{r}(s,s')} \Big] \\
    & \hspace{30pt} + \E_{(s,a) \sim \blue{d}} \Big[ \underbrace{\E_{s'\sim\hat{T}(\cdot|s,a)}[-\mu(s,s') + \gamma \nu(s')] - \nu(s)}_{=:\widehat{e}^I_{\mu,\nu}(s,a)} - \alpha\log\underbrace{\tfrac{\blue{d}(s,a)}{\widehat{d}^I(s,a)}}_{=:\blue{w}(s,a)} \Big]\\
    =& \min_{\mu,\nu} \max_{\blue{w},\redb{w}\ge0} (1 - \gamma) \E_{s_0 \sim \widehat{p}_0} [\nu(s_0)] + \E_{(s,s') \sim \widehat{d}^I} \big[ \redb{w}(s,s') \big(\widehat r(s,s') + \mu(s,s') - \log\redb{w}(s,s') \big)\big] \\
    & \hspace{30pt} + \E_{(s,a) \sim \widehat d^I} \big[\blue{w}(s,a) \big( \widehat{e}^I_{\mu,\nu}(s,a) - \alpha\log\blue{w}(s,a) \big)\big] =: \gL^I(\blue{w},\redb{w},\mu,\nu),
\end{align*}
Note that this is identical to \eqref{eq:maximin_lagrangian_detail_sample}. We can also derive the closed-form solution $(\blue{w}^I_{\mu,\nu},\redb{w}^I_{\mu})$ to the $\max_{\blue{w},\redb{w}\ge0}\gL^I(\blue{w},\redb{w},\mu,\nu)$:
\begin{align}
    \blue{w}^I_{\mu,\nu}(s,a)&=\exp\Big(\tfrac{1}{\alpha} \widehat{e}^I_{\mu,\nu}(s,a) -1\Big)
    \emph{ ~and~ }
    \redb{w}^I_{\mu}(s,s')=\exp(\widehat r(s,s')+\mu(s,s')-1).
    \label{eq:closed_form_solution_wss1_mlemdp}
\end{align}
Note that this is different from \eqref{eq:closed_form_solution_wss1_sample} in that \eqref{eq:closed_form_solution_wss1_mlemdp} is the closed-form solution for each $(s,a)$ and $(s,s')$, while \eqref{eq:closed_form_solution_wss1_sample} is the \emph{non-parametric} closed-form solution for each sample $x \in D^I$.
Exploiting \eqref{eq:closed_form_solution_wss1_mlemdp} does \emph{not} introduce an additional bias, but requires evaluation of the expectation within the $\exp(\cdot)$, i.e. transition model is needed.

The last step is to plug the solution $(\blue{w}^I_{\mu,\nu},\redb{w}^I_{\mu})$ in \eqref{eq:closed_form_solution_wss1_mlemdp} into $\gL^I(\blue{w},\redb{w},\mu,\nu)$:
\begin{align*}
    \min_{\mu,\nu}\, & \gL^I (\blue{w}^I_{\mu,\nu},\redb{w}^I_{\mu},\mu, \nu) 
    = (1-\gamma)\E_{s \sim \widehat{p}_0}[\nu(s)] + \E_{(s,s') \sim \widehat{d}^I} \Big[ \exp\big(\widehat r(s,s')+\mu(s,s')-1\big) \Big] \\
    &  + \alpha\E_{(s,a) \sim \widehat d^I}\Big[ \exp\big(\tfrac{1}{\alpha} \widehat{e}^I_{\mu,\nu}(s,a) -1\big)\Big].
\end{align*}
Finally, this can be rewritten as:
\begin{align}
    \min_{\mu,\nu}\, &
    \textstyle J (\mu, \nu) 
    = (1-\gamma) \sum\limits_{s} \widehat p_0(s) \nu(s) + \sum\limits_{s,s'} \widehat{d}^I(s,s') \Big[ \exp\big(\widehat r(s,s') + \mu(s,s')-1\big) \Big] \nonumber \\
    & \textstyle  + \alpha \sum\limits_{s,a} \widehat d^I(s,a) \Big[ \exp \Big( \tfrac{1}{\alpha} \sum\limits_{s'} \widehat{T}^I(s'|s,a) \big( -\mu(s,s') + \gamma \nu(s') - \nu(s) \big) - 1 \Big) \Big], \nonumber
\end{align}
which is the objective function for our tabular LobsDICE~\eqref{eq:tabular_lobsdice_objective}.

\subsection{When will LobsDICE reduce to BCO?}
LobsDICE reduces to BCO when $D^I$ is collected by an \emph{expert} policy and $D^E$ is a dataset identical to $D^I$ except that action is missing, i.e., $D^E=\{(s,s')|(s,a,s')\in D^I\}$.

In this situation, note that $\red{\widehat{d}^I}(s,s')$ and $\blue{\widehat{d}^I}(s,a)$ are valid stationary distributions on the MLE MDP constructed by $D^I$, since they satisfy both the Bellman flow constraint \eqref{eq:sample_based_const1} and the marginalization constraint \eqref{eq:sample_based_const2} on the MLE MDP.
The $\red{\widehat{d}^I}(s,s')$ and $\blue{\widehat{d}^I}(s,a)$ are also the optimal solution of (\ref{eq:sample_based_obj}-\ref{eq:sample_based_const2}), making the KL divergences to 0: 
\begin{align*}
    \KL(\red{\widehat{d}^I}(s,s') \| \widehat{d}^E(s,s') ) = 0 \text{ and } \KL(\blue{\widehat{d}^I}(s,a) \| \widehat{d}^I(s,a)) = 0.
\end{align*}
Finally, the policy obtained by LobsDICE will simply reduce to BC on $D^I$ by noting that:
\begin{align*}
    \pi^*(a|s) = \frac{\blue{\widehat{d}^I}(s,a)}{\sum_{a'}\blue{\widehat{d}^I}(s,a')}
\end{align*}

Furthermore, one can easily show that BC on $D^I$ is identical to BCO (i.e. BC on $D^E$ where the missing actions are inferred by the IDM trained by $D^I$), which concludes that LobsDICE is equivalent to BCO in this special case.


However, $D^I$ will include demonstrations collected by \emph{non-expert} policies in general, which makes the LobsDICE's solution different from (and usually better than) the BCO's solution in general cases.

\section{Theoretical Analysis}
\label{app:theoretical_analysis}
\subsection{Closed-form solutions}
\firstclosedformsolution*
\begin{proof}
Let $e_{\mu,\nu}(s,a,s')=-\mu(s,s')+\gamma\nu(s')-\nu(s)$. Then, $e_{\mu,\nu}(s,a)=\E_{s'\sim T(\cdot|s,a)}[e_{\mu,\nu}(s,a,s')]$.
For $(s,a)$ with $d^I(s,a)>0$, we can compute the derivative $\frac{\partial\gL}{\partial\blue{w}(s,a)}$ of $\gL$ w.r.t. $\blue{w}(s,a)$ as follows:
\begin{align*}
    \frac{\partial\gL}{\partial\blue{w}(s,a)}
    &=\sum_{s'}d^I(s,a,s')(e_{\mu,\nu}(s,a,s')-\alpha\log\blue{w}(s,a)-\alpha)=0\\
    &\Leftrightarrow\sum_{s'}T(s'|s,a)(e_{\mu,\nu}(s,a,s')-\alpha\log\blue{w}(s,a)-\alpha)=0\\
    &\Leftrightarrow\sum_{s'}T(s'|s,a)(e_{\mu,\nu}(s,a,s')-\alpha)=\alpha\log\blue{w}(s,a)\\
    &\Leftrightarrow\blue{w}(s,a)=\exp\Big(\tfrac{1}{\alpha}\E_{s'\sim T(\cdot|s,a)}[e_{\mu,\nu}(s,a,s')]-1\Big)
    =\exp\Big(\tfrac{1}{\alpha}e_{\mu,\nu}(s,a)-1\Big).
\end{align*}
Similar to the aforementioned derivation, when $\bar{d}^I(s,s')>0$, we can derive the derivative of $\gL$ w.r.t. $\redb{w}(s,s')$:
\begin{align*}
    \frac{\partial\gL}{\partial\redb{w}(s,s')}
    &=\bar{d}^I(s,s')\big(\mu(s,s')-\log\redb{w}(s,s')+r(s,s')-1\big)=0\\
    &\Leftrightarrow\big(\mu(s,s')-\log\redb{w}(s,s')+r(s,s')-1\big)=0\\
    &\Leftrightarrow\log\redb{w}(s,s')=\mu(s,s')+r(s,s')-1\\
    &\Leftrightarrow\redb{w}(s,s')=\exp\big(\mu(s,s')+r(s,s')-1\big).
\end{align*}
\end{proof}
\convexitydouble*
\begin{proof}
Using the fact that $\exp(\cdot)$ is a convex function, we can easily prove the convexity of $\gL(\blue{w}_{\mu,\nu},\redb{w}_{\mu,\nu},\mu,\nu)$.
For brevity, let $\gL(\mu,\nu):=\gL(\blue{w}_{\mu,\nu},\redb{w}_{\mu,\nu},\mu,\nu)$.
Then, for given $(\mu_1,\nu_1)$, $(\mu_2,\nu_2)$ and $t\in[0,1]$,
\begin{align*}
    \gL~&(t\mu_1+(1-t)\mu_2,t\nu_1+(1-t)\nu_2)\\
    =~&(1-\gamma)\E_{p_0}[t\nu_1(s)+(1-t)\nu_2(s)] + \E_{d^I}[\exp(r(s,s')+t\mu_1(s,s')+(1-t)\mu_2(s,s')-1)] \\
    & + \alpha\E_{d^I}\Big[\exp\Big(\tfrac{1}{\alpha}\E_{s'}\big[-t\mu_1(s,s')-(1-t)\mu_2(s,s')+\gamma t\nu_1(s')+\gamma(1-t)\nu_2(s')\\
    &\hspace{3.2cm}-t\nu_1(s)-(1-t)\nu_2(s)\big]-1\Big)\Big]\\
    =~& t(1-\gamma)\E_{p_0}[\nu_1(s)]+(1-t)(1-\gamma)\E_{p_0}[\nu_2(s)] + \E_{d^I}[\exp(r(s,s')+t\mu_1+(1-t)\mu_2-1)] \\
    & + \alpha\E_{d^I}\Big[\exp\Big(\tfrac{1}{\alpha}\E_{s'}\big[-t\mu_1(s,s')-(1-t)\mu_2(s,s')+\gamma t\nu_1(s')+\gamma(1-t)\nu_2(s')\\
    &\hspace{3.2cm}-t\nu_1(s)-(1-t)\nu_2(s)\big]-1\Big)\Big]
\end{align*}
\begin{align*}
    (Cont.)\le~& t(1-\gamma)\E_{p_0}[\nu_1(s)]+(1-t)(1-\gamma)\E_{p_0}[\nu_2(s)]\\
    & + t\E_{d^I}[\exp(r(s,s')+\mu_1(s,s')-1)]+(1-t)\E_{d^I}[\exp(r(s,s')+\mu_2(s,s')-1)] \\
    & + \alpha\E_{d^I}\Big[\exp\Big(\tfrac{1}{\alpha}\E_{s'}\big[-t\mu_1(s,s')-(1-t)\mu_2(s,s')+\gamma t\nu_1(s')+\gamma(1-t)\nu_2(s')\\
    &\hspace{3.2cm}-t\nu_1(s)-(1-t)\nu_2(s)\big]-1\Big)\Big]\\
    \le~& t(1-\gamma)\E_{p_0}[\nu_1(s)]+(1-t)(1-\gamma)\E_{p_0}[\nu_2(s)]\\
    & + t\E_{d^I}[\exp(r(s,s')+\mu_1(s,s')-1)]+(1-t)\E_{d^I}[\exp(r(s,s')+\mu_2(s,s')-1)] \\
    & + t\alpha\E_{d^I}\Big[\exp\Big(\tfrac{1}{\alpha}\E_{s'}\big[\mu_1(s,s')+\gamma \nu_1(s')-\nu_1(s)\big]-1\Big)\Big]\\
    &+ (1-t) \alpha\E_{d^I}\Big[\exp\Big(\tfrac{1}{\alpha}\E_{s'}\big[-\mu_2(s,s')+\gamma\nu_2(s')-\nu_2(s)\big]-1\Big)\Big]\\
    =~&t\gL(\mu_1,\nu_1)+(1-t)\gL(\mu_2,\nu_2).
\end{align*}
For the inequalities in the above formulation, we use the fact that $\E_{d^I}[\exp(\cdot)]$ is a instance of convex functions.
\end{proof}
\secondclosedformsolution*
\begin{proof}
For simplicity, let $\widehat\gL (\mu, \nu) := \widehat\gL (\blue{w}_{\mu,\nu},\redb{w}_{\mu,\nu},\mu, \nu)$ and $x:=(s,a,s')$. Then,
\begin{align*}
    \tfrac{\partial\widehat\gL}{\partial\mu(x)}
    &=d^I(x)\Big[\exp(r(x)+\mu(x)-1)-\exp\Big(\tfrac{1}{\alpha}(-\mu(x)+\gamma\nu(s')-\nu(s))-1\Big)\Big]=0\\
    &\Leftrightarrow\exp(r(x)+\mu(x)-1)=\exp\Big(\tfrac{1}{\alpha}(-\mu(x)+\gamma\nu(s')-\nu(s))-1\Big)\\
    &\Leftrightarrow \alpha(r(x)+\mu(x))=-\mu(x)+\gamma\nu(s')-\nu(s)\\
    &\Leftrightarrow (1+\alpha)\mu(x)=-\alpha r(x)+\gamma\nu(s')-\nu(s)\\
    &\Leftrightarrow \mu(x)=\tfrac{1}{1+\alpha}\big(-\alpha r(x)+\gamma\nu(s')-\nu(s)\big)
\end{align*}
\end{proof}

\subsection{Surrogate Objective}
We first show a property of $\tilde{\gL}$ that will be used to prove propositions:
\begin{lemma}
For given function $\nu:S\rightarrow\R$,
\[
\tilde{\gL}(\nu)=\tilde{\gL}(\nu+C)\quad\forall C\in\R.
\]
\label{lem:fd_const_invariance}
\end{lemma}
\begin{proof}
For any $C\in\R$,
\begin{align*}
    &\tilde{\gL}(\nu+C)\\
    &=(1-\gamma)\E_{s\sim p_0}[\nu(s)+C]+(1+\alpha)\log\E_{(s,a,s')\sim d^I}\Big[\exp\Big(\tfrac{1}{1+\alpha}\hat{A}_{\nu+C}(s,a,s')\Big)\Big]\\
    &=(1-\gamma)\E_{s\sim p_0}[\nu(s)+C]+(1+\alpha)\log\E_{(s,a,s')\sim d^I}\Big[\exp\Big(\tfrac{1}{1+\alpha}\big(r(s,s')+\gamma(\nu(s')+C)-(\nu(s)+C)\big)\Big)\Big]\\
    &=(1-\gamma)\E_{s\sim p_0}[\nu(s)+C]+(1+\alpha)\log\E_{(s,a,s')\sim d^I}\Big[\exp\Big(\tfrac{1}{1+\alpha}\big(r(s,s')+\gamma\nu(s')-\nu(s)\big)\Big)\exp\Big(\tfrac{(\gamma-1)C}{1+\alpha}\Big)\Big]\\
    &=(1-\gamma)\E_{s\sim p_0}[\nu(s)+C]+(1+\alpha)\log\E_{(s,a,s')\sim d^I}\Big[\exp\Big(\tfrac{1}{1+\alpha}(r(s,s')+\gamma\nu(s')-\nu(s))\Big)\Big]+(\gamma-1)C\\
    &=(1-\gamma)\E_{s\sim p_0}[\nu(s)]+(1+\alpha)\log\E_{(s,a,s')\sim d^I}\Big[\exp\Big(\tfrac{1}{1+\alpha}(r(s,s')+\gamma\nu(s')-\nu(s))\Big)\Big]\\
    &=\tilde{\gL}(\nu).
\end{align*}
\end{proof}
Now, we prove Proposition~\ref{prop:fencheldual} and Proposition~\ref{prop:ld_fd_connection}:
\fencheldual*
\begin{proof}
First of all, from the fact that $\log (x+1)\le x$ for all $x>-1$, we can easily conclude that $\min_\nu\tilde{\gL}(\nu)\le\min_\nu\hat\gL(\nu)$.
Now, we will show that $\min_\nu\hat\gL(\nu)\le\min_\nu\tilde{\gL}(\nu)$.
For given $\nu^*\in\argmin_{\nu}\tilde{\gL}(\nu)$, we define a constant $C$ as follows:
\begin{align*}
    C&=\tfrac{1+\alpha}{1-\gamma}\log\E_{(s,a,s')\sim d^I}\Big[\exp\Big(\tfrac{1}{1+\alpha}\hat{A}_{\nu^*}(s,a,s')-1\Big)\Big],
\end{align*}
which implies 
\[
\exp\Big(\tfrac{(1-\gamma)C}{1+\alpha}\bigg)=\E_{(s,a,s')\sim d^I}\Big[\exp\Big(\tfrac{1}{1+\alpha}\hat{A}_{\nu^*}(s,a,s')-1\Big)\Big].
\]
Then, from the following two equations:
\begin{align*}
    &\hat\gL(\nu^*+C)\\
    &=(1-\gamma)\E_{s\sim p_0}[\nu^*(s)+C]+(1+\alpha)\E_{(s,a,s')\sim d^I}\Big[\exp\Big(\tfrac{1}{1+\alpha}\hat{A}_{\nu^*+C}(s,a,s')-1\Big)\Big]\\
    &=(1-\gamma)\E_{s\sim p_0}[\nu^*(s)+C]+(1+\alpha)\E_{(s,a,s')\sim d^I}\Big[\exp\Big(\tfrac{1}{1+\alpha}\hat{A}_{\nu^*}(s,a,s')-1\Big)\exp\Big(\tfrac{(\gamma-1)C}{1+\alpha}\Big)\Big]\\
    &=(1-\gamma)\E_{s\sim p_0}[\nu^*(s)+C]+(1+\alpha)\E_{(s,a,s')\sim d^I}\Big[\exp\Big(\tfrac{1}{1+\alpha}\hat{A}_{\nu^*}(s,a,s')-1\Big)\Big]\exp\Big(-\tfrac{(1-\gamma)C}{1+\alpha}\Big)\\
    &=(1-\gamma)\E_{s\sim p_0}[\nu^*(s)+C]+(1+\alpha)
\end{align*}
and
\begin{align*}
    &\tilde{\gL}(\nu^*+C)\\
    &=(1-\gamma)\E_{s\sim p_0}[\nu^*(s)+C]+(1+\alpha)\log\E_{(s,a,s')\sim d^I}\Big[\exp\Big(\tfrac{1}{1+\alpha}\hat{A}_{\nu^*+C}(s,a,s')\Big)\Big]\\
    &=(1-\gamma)\E_{s\sim p_0}[\nu^*(s)+C]+(1+\alpha)\log\E_{(s,a,s')\sim d^I}\Big[\exp\Big(\tfrac{1}{1+\alpha}\hat{A}_{\nu^*}(s,a,s')\Big)\Big]+(\gamma-1)C\\
    &=(1-\gamma)\E_{s\sim p_0}[\nu^*(s)+C]+(1+\alpha)\log\E_{(s,a,s')\sim d^I}\Big[\exp\Big(\tfrac{1}{1+\alpha}\hat{A}_{\nu^*}(s,a,s')\Big)\Big]\\
    &\quad-(1+\alpha)\log\E_{(s,a,s')\sim d^I}\Big[\exp\Big(\tfrac{1}{1+\alpha}\hat{A}_{\nu^*}(s,a,s')-1\Big)\Big]\\
    &=(1-\gamma)\E_{s\sim p_0}[\nu^*(s)+C]+(1+\alpha)\log\E_{(s,a,s')\sim d^I}\Big[\exp\Big(\tfrac{1}{1+\alpha}\hat{A}_{\nu^*}(s,a,s')\Big)\Big]\\
    &\quad-(1+\alpha)\Big(\log\E_{(s,a,s')\sim d^I}\Big[\exp\Big(\tfrac{1}{1+\alpha}\hat{A}_{\nu^*}(s,a,s')\Big)\Big]-1\Big)\\
    &=(1-\gamma)\E_{s\sim p_0}[\nu^*(s)+C]+1+\alpha,
\end{align*}
we can conclude that 
\[
\hat\gL(\nu^*+C) = \tilde{\gL}(\nu^*+C).
\]
From the Lemma~\ref{lem:fd_const_invariance}, we obtain
\[
\min_{\nu}\hat\gL(\nu)\le\hat\gL(\nu^*+C) = \tilde{\gL}(\nu^*+C) = \min_{\nu}\tilde{\gL}(\nu).
\]
We show that $\min_\nu\tilde{\gL}(\nu)\le\min_\nu\hat\gL(\nu)$ and $\min_\nu\hat\gL(\nu)\le\min_\nu\tilde{\gL}(\nu)$, so $\min_{\nu}\hat\gL(\nu)=\min_{\nu}\tilde{\gL}(\nu)$.

Finally, similar to the proof steps for Proposition~\ref{prop:convexity_double}, we can easily show that $\hat\gL_\mathrm{FD}(\nu)$ is convex w.r.t. $\nu$ (Remark that log-sum-exp is a convex function).
\end{proof}


\ldfdconnection*
\begin{proof}
We will prove this proposition by showing $\tilde{V}\subseteq \{\nu^*+C'|\nu^*\in \hat{V}, C\in\R\}$ and $\tilde{V}\supseteq \{\nu^*+C'|\nu^*\in \hat{V}, C\in\R\}$.
\\
$(\subseteq)$
For given $\nu^*\in\argmin_{\nu}\tilde{\gL}(\nu)$, we define a constant $C$ as follows:
\begin{align*}
    C&=\tfrac{1+\alpha}{1-\gamma}\log\E_{(s,a,s')\sim d^I}\Big[\exp\Big(\tfrac{1}{1+\alpha}\hat{A}_{\nu^*}(s,a,s')-1\Big)\Big],
\end{align*}
which implies 
\[
\exp\Big(\tfrac{(1-\gamma)C}{1+\alpha}\Big)=\E_{(s,a,s')\sim d^I}\Big[\exp\Big(\tfrac{1}{1+\alpha}\hat{A}_{\nu^*}(s,a,s')-1\Big)\Big].
\]
Then, by the proof steps of Proposition~\ref{prop:fencheldual},
we obtain 
\[
\hat\gL(\nu^*+C) = \tilde{\gL}(\nu^*+C) = \min_{\nu}\tilde{\gL}(\nu) = \min_{\nu}\hat\gL(\nu).
\]
It means $(\nu^*+C)\in\argmin_{\nu}\hat\gL(\nu)$ and thus, 
\[
\nu^*\in\{\hat\nu^*+C|\hat\nu^*\in\argmin_{\nu}\hat\gL(\nu), C\in\R\},
\]
i.e., 
\[
\argmin_{\nu}\tilde{\gL}(\nu)\subseteq\{\hat\nu^*+C|\hat\nu^*\in\argmin_{\nu}\hat\gL(\nu), C\in\R\}
\]

$(\supseteq)$
Let $\hat\nu^*\in\argmin_{\nu}\hat\gL(\nu)$. Then, 
\begin{align*}
    \frac{d^*(s,a)}{d^I(s,a)}=w^*_{\mu^*,\nu^*}(s,a)
    &=\exp\Big(\tfrac{1}{\alpha}\big(\E_{s'\sim T(\cdot|s,a)} [\gamma\nu(s')-\mu(s,s')]-\nu(s)\big)-1\Big),\\
    \frac{d^*(s,s')}{d^I(s,s')}=\bar{w}^*_{\mu^*,\nu^*}(s,s')
    &=\exp(r(s,s')+\mu^*(s,s')-1).
\end{align*}

Let $C\in\R$. Then, we derive the following equation:
\begin{align*}
    &\tilde{\gL}(\hat\nu^*+C)=\tilde{\gL}(\hat\nu^*)\\
    &=(1-\gamma)\E_{s\sim p_0}[\hat\nu^*(s)]+(1+\alpha)\log\E_{(s,a,s')\sim d^I}\Big[\exp\Big(\tfrac{1}{1+\alpha}\hat{A}_{\hat\nu^*}(s,a,s')\Big)\Big]\\
    &=(1-\gamma)\E_{s\sim p_0}[\hat\nu^*(s)]+(1+\alpha)\log\E_{(s,a,s')\sim d^I}\Big[\exp\Big(\tfrac{1}{1+\alpha}\hat{A}_{\hat\nu^*}(s,a,s')-1+1\Big)\Big]\\
    &=(1-\gamma)\E_{s\sim p_0}[\nu^*(s)]+1+\alpha.
\end{align*}
Here, we apply Lemma~\ref{lem:fd_const_invariance} to derive the first equality. Because
\begin{align*}
    \min_{\nu}\tilde{\gL}(\nu)
    &=\min_{\nu}\hat\gL(\nu)\\
    &=\hat\gL(\hat\nu^*)\\
    &=(1-\gamma)\E_{s\sim p_0}[\hat\nu^*(s)]+(1+\alpha)\E_{(s,a,s')\sim d^I}\Big[\exp\Big(\tfrac{1}{1+\alpha}\hat{A}_{\hat\nu^*}(s,a,s')-1\Big)\Big]\\
    &=(1-\gamma)\E_{s\sim p_0}[\nu^*(s)]+1+\alpha,
\end{align*}
we can conclude that $\gL_\text{FD}(\hat\nu^*+C)=\min_{\nu}\gL_\text{FD}(\nu)$, i.e., $(\hat\nu^*+C)\in\argmin_{\nu}\tilde{\gL}(\nu)$. Consequently,
\[
\argmin_{\nu}\tilde{\gL}(\nu)\supseteq\{\hat\nu^*+C|\hat\nu^*\in\argmin_{\nu}\hat\gL(\nu), C\in\R\}.
\]
\end{proof}
\subsection{Fenchel dual formulation}
Let 
\[
\delta_C(x):=
    \begin{cases}
    0&x\in C\\
    \infty&\text{otherwise}
    \end{cases}.
\]
Then we can provide following proposition:
\begin{proposition}
We can rewrite the optimization problem (\ref{eq:main_objective}-\ref{eq:marginalization_constraint}) as
\begin{align}
    \max_{d\ge0}
    & - \delta_{(1-\gamma)p_0} (-(\gamma\gT-\gB)_*d) - \KL\big((\bar{\gT}_*d)\|\bar{d^E}\big) - \alpha\KL(d\|d^I) .
    \label{eq:primal_fenchel}
\end{align}
Then, the dual problem of \eqref{eq:primal_fenchel} is given by 
\begin{align}
    &\min_{\mu,\nu}  \gL_\text{FD} (\mu, \nu)\label{eq:dual_fenchel}\\
    &:= (1-\gamma)\E_{s \sim p_0}[\nu(s)]  + \log\E_{\bar{d}^I}\big[\exp\big(r(s,s')+\mu(s,s')\big)\big]
    + \alpha\log\E_{d^I}\Big[\exp\Big(\tfrac{1}{\alpha} e_{\mu,\nu}(s,a)\Big)\Big].\nonumber
\end{align}
\end{proposition}
\begin{proof}
We first define following three functions
\begin{align*}
    f(\cdot)&:= \delta_{\{(1-\gamma)p_0\}}(\cdot),\\
    g(\cdot;r)&:= \langle \cdot, -r \rangle+\KL(\cdot\|\bar\gT_*d^I),\\
    h(\cdot;\mu)&:=\langle \cdot, \bar\gT\mu \rangle + \alpha\KL(\cdot\|d^I),
\end{align*}
and conjugate functions,
\begin{align*}
    f_*(\cdot)&:=(1-\gamma)\E_{s\sim p_0}[\cdot],\\
    g_*(\cdot;r)&:=\log\E_{(s,s')\sim\bar\gT_*d^I}[\exp(\cdot+r(s,s'))],\\
    h_*(\cdot;\mu)&:=\alpha\log\E_{(s,a)\sim d^I}\bigg[\exp\bigg(\frac{\cdot-(\bar\gT\mu)(s,a)}{\alpha}\bigg)\bigg].
\end{align*}
Then, the dual formulation of the primal (\ref{eq:primal_fenchel}) can be derived as follows:
\begin{align*}
    \max_{d\ge0}&-\delta_{(1-\gamma)p_0}\big(-(\gamma\gT-\gB)_*d\big) -\E_{\substack{(s,a)\sim d,\\s'\sim T(\cdot|s,a)}}\bigg[\log\frac{(\bar{\gT}_*d)(s,s')}{d^E(s,s')}\bigg]-\alpha\KL(d\|d^I)\\
    =\max_{d\ge0}&-f\big(-(\gamma\gT-\gB)_*d\big) +\E_{\substack{(s,a)\sim d,\\s'\sim T(\cdot|s,a)}}\bigg[-\log\frac{(\bar{\gT}_*d)(s,s')}{(\bar\gT_*d^I)(s,s')}+\underbrace{\log\frac{d^E(s,s')}{(\bar\gT_*d^I)(s,s')}}_{:=r(s,s')}\bigg]-\alpha\KL(d\|d^I)\\
    =\max_{d\ge0}&-f\big(-(\gamma\gT-\gB)_*d\big) - g(\bar\gT_*d;r)-\alpha\KL(d\|d^I)\\
    =\max_{d\ge0}& - f\big(-(\gamma\gT-\gB)_*d\big) - \bigg\{\max_\mu-g_*(\mu;r)+\langle\bar\gT_*d,\mu\rangle\bigg\}  -\alpha\KL(d\|d^I)\\
    =\max_{d\ge0}\min_\mu& - f\big(-(\gamma\gT-\gB)_*d\big) + g_*(\mu;r) - \langle\bar\gT_*d,\mu\rangle  -\alpha\KL(d\|d^I) \\
    =\max_{d\ge0}\min_\mu& - f\big(-(\gamma\gT-\gB)_*d\big) + g_*(\mu;r) - \langle d,\bar\gT\mu\rangle  -\alpha\KL(d\|d^I)\\
    =\max_{d\ge0}\min_\mu&  - f\big(-(\gamma\gT-\gB)_*d\big) + g_*(\mu;r) - h(d;\mu)\\
    =\max_{d\ge0}\min_\mu&  - \bigg\{\max_\nu\langle-(\gamma\gT-\gB)_*d,\nu\rangle -f_*(\nu)\bigg\}+ g_*(\mu;r) - h(d;\mu)\\
    =\max_{d\ge0}\min_{\nu,\mu}& \langle (\gamma\gT-\gB)_*d,\nu\rangle  + f_*(\nu) + g_*(\mu;r) - h(d;\mu)\\
    =\max_{d\ge0}\min_{\nu,\mu}& \langle d,(\gamma\gT-\gB)\nu\rangle  + f_*(\nu) + g_*(\mu;r) - h(d;\mu):=\gL_{\text{FD}}(d;\mu,\nu)\\
\end{align*}
Here, we can reorder the maximin to minimax and therefore, we can derive the 
\begin{align*}
    \min_{\nu,\mu}\max_{d\ge0}&\gL_{\text{FD}}(d;\mu,\nu)\\
    =\min_{\nu,\mu}\max_{d\ge0}&\langle d,(\gamma\gT-\gB)\nu\rangle  + f_*(\nu) + g_*(\mu;r) - h(d;\mu)\\
    =\min_{\nu,\mu}&\bigg\{\max_{d\ge0}\langle d,(\gamma\gT-\gB)\nu\rangle - h(d;\mu)\bigg\} + f_*(\nu) + g_*(\mu;r) \\
    =\min_{\nu,\mu}& h_*((\gamma\gT-\gB)\nu) + f_*(\nu) + g_*(\mu;r).
\end{align*}

We can rewrite the last term as
\begin{align*}
    \min_{\nu,\mu}& (1-\gamma)\E_{s\sim p_0}[\nu(s)] + 
    \log\E_{\substack{(s,a)\sim d^I,\\s'\sim T(\cdot|s,a)}}[\exp(\mu(s,s')+r(s,s'))]
    + \alpha\log\E_{(s,a)\sim d^I}\Big[\exp\Big(\tfrac{1}{\alpha}e_{\mu,\nu}(s,a)\Big)\Big].
\end{align*}
\end{proof}

\begin{lemma}
For given function $\mu:S\times S\rightarrow\R$ and $\nu:S\rightarrow\R$,
\[
\gL_\text{FD}(\mu,\nu)=\gL_\text{FD}(\mu+C,\nu+C')\quad\forall C,C'\in\R.
\]
\end{lemma}
\begin{proof}
\begin{align*}
    &\gL_\text{FD}(\mu+C,\nu+C')\\
    &=(1-\gamma)\E_{s\sim p_0}[\nu(s)+C']+\log\E_{(s,a,s')\sim d^I}[\exp(\mu(s,s')+C+r(s,s'))]\\
    &\quad+\alpha\log\E_{(s,a)\sim d^I}\Big[\exp\Big(\tfrac{1}{\alpha} \big(\E_{s'\sim T(\cdot|s,a)}[\gamma\nu(s')+\gamma C'-\mu(s,s')-C]-\nu(s)-C'\big) \Big)\Big]\\
    &=(1-\gamma)\E_{s\sim p_0}[\nu(s)]+(1-\gamma)C'+\log\E_{(s,a,s')\sim d^I}[\exp(\mu(s,s')+r(s,s'))]+C\\
    &\quad+\alpha\log\E_{(s,a)\sim d^I}\Big[\exp\Big(\tfrac{1}{\alpha} \big( \E_{s'\sim T(\cdot|s,a)}[\gamma\nu(s')-\mu(s,s')]-\nu(s) \big) \Big)\Big]+(\gamma C'-C-C')\\
    &=(1-\gamma)\E_{s\sim p_0}[\nu(s)]+\log\E_{(s,a,s')\sim d^I}[\exp(\mu(s,s')+r(s,s'))]\\
    &\quad+\alpha\log\E_{(s,a)\sim d^I}\Big[\exp\Big(\tfrac{1}{\alpha} \big(\E_{s'\sim T(\cdot|s,a)}[\gamma\nu(s')-\mu(s,s')]-\nu(s) \big) \Big)\Big]\\
    &=\gL_\text{FD}(\mu,\nu).
\end{align*}
\end{proof}

Finally, we can show that the relation between $\argmin_{\mu,\nu}\gL$ and $\argmin_{\mu,\nu}\gL_\mathrm{FD}$:

\begin{proposition}
Let $V$ and $V_\text{FD}$ be the set of optimal solutions $(\mu^*,\nu^*)$ of $\argmin_{\mu,\nu}\gL(\blue{w}_{\nu,\mu},\redb{w}_{\nu,\mu},\mu,\nu)$ and $\argmin_{\mu,\nu}\gL_\mathrm{FD}(\mu,\nu)$, respectively. Then,
\begin{equation*}
    V_\text{FD}=\{(\mu^*+C,\nu^*+C')|(\mu^*,\nu^*)\in V, C\in\R,C'\in\R\}
\end{equation*}
holds.
\end{proposition}
\begin{proof}
For brevity, we will denote $\gL(\blue{w}_{\nu,\mu},\redb{w}_{\nu,\mu},\mu,\nu)$ as $\gL(\mu,\nu)$.

$(\subseteq)$
For given $(\hat\mu^*,\hat\nu^*)\in\argmin_{\mu,\nu}\gL_\text{FD}(\mu,\nu)$, we define two constants as follows:
\begin{align*}
    C&=-\log\E_{(s,a,s')\sim d^I}[\exp(\hat\mu^*(s,s')+r(s,s')-1)],\\
    C'&=\frac{\alpha}{1-\gamma}\log\E_{(s,a)\sim d^I}\Big[\exp\Big(\frac{\E_{s'}[\gamma\hat\nu^*(s')-\hat\mu^*(s,s')]-\hat\nu^*(s')}{\alpha}-1\Big)\Big]-\frac{C}{1-\gamma}.
\end{align*}
Then, from the following two equations:
\begin{align*}
    &\gL(\hat\mu^*+C,\hat\nu^*+C')\\
    &=(1-\gamma)\E_{s\sim p_0}[\hat\nu^*(s)+C']+\E_{(s,a,s')\sim d^I}[\exp(\hat\mu^*(s,s')+C+r(s,s')-1)]\\
    &\quad+\alpha\E_{(s,a)\sim d^I}\Big[\exp\Big(\tfrac{1}{\alpha} \big( \E_{s'}[\gamma\hat\nu^*(s')+\gamma C'-\hat\mu^*(s,s')-C]-\hat\nu^*(s')-C' \big) -1\Big)\Big]\\
    &=(1-\gamma)\E_{s\sim p_0}[\hat\nu^*(s)]+(1-\gamma)C'+\E_{(s,a,s')\sim d^I}[\exp(\hat\mu^*(s,s')+r(s,s')-1)]\exp(C)\\
    &\quad+\alpha\E_{(s,a)\sim d^I}\Big[\exp\Big(\tfrac{1}{\alpha}\big( \E_{s'}[\gamma\hat\nu^*(s')-\hat\mu^*(s,s')]-\hat\nu^*(s') \big) -1\Big)\Big]\exp\Big(\tfrac{1}{\alpha} \big(-(1 - \gamma) C'-C \big) \Big)\\
    &=(1-\gamma)\E_{s\sim p_0}[\hat\nu^*(s)]+(1-\gamma)C'+1+\alpha,
\end{align*}
and
\begin{align*}
    &\gL_\text{FD}(\hat\mu^*+C,\hat\nu^*+C')\\
    &=(1-\gamma)\E_{s\sim p_0}[\hat\nu^*(s)+C']+\log\E_{(s,a,s')\sim d^I}[\exp(\hat\mu^*(s,s')+C+r(s,s'))]\\
    &\quad+\alpha\log\E_{(s,a)\sim d^I}\Big[\exp\Big(\tfrac{1}{\alpha} \big( \E_{s'}[\gamma\hat\nu^*(s')+\gamma C'-\hat\mu^*(s,s')-C]-\hat\nu^*(s')-C'\big) \Big)\Big]\\
    &=(1-\gamma)\E_{s\sim p_0}[\hat\nu^*(s)]+(1-\gamma)C'+\log\E_{(s,a,s')\sim d^I}[\exp(\hat\mu^*(s,s')+r(s,s'))]+C\\
    &\quad+\alpha\log\E_{(s,a)\sim d^I}\Big[\exp\Big(\tfrac{1}{\alpha} \big( \E_{s'}[\gamma\hat\nu^*(s')-\hat\mu^*(s,s')]-\hat\nu^*(s')\big) \Big)\Big]+(\gamma C'-C-C')\\
    &=(1-\gamma)\E_{s\sim p_0}[\hat\nu^*(s)+C']+1+\alpha,
\end{align*}
we can conclude that 
\[
\gL(\hat\mu^*+C,\hat\nu^*+C') = \gL_\text{FD}(\hat\mu^*+C,\hat\nu^*+C') = \gL_\text{FD}(\hat\mu^*,\hat\nu^*)=\min_{\mu,\nu}\gL_\text{FD}(\mu,\nu) = \min_{\mu,\nu}\gL(\mu,\nu).
\]
It means $(\hat\mu^*+C,\hat\nu^*+C')\in\argmin_{\mu,\nu}\gL(\mu,\nu)$ and thus, 
\[
(\hat\mu^*,\hat\nu^*)\in\{\mu+C,\nu+C'|(\mu,\nu)\in\argmin_{\mu,\nu}\gL(\mu,\nu), C\in\R,C'\in\R\},
\]
i.e., 
\[
\argmin_{\mu,\nu}\gL_\text{FD}(\mu,\nu)\subseteq\{\mu+C,\nu+C'|(\mu,\nu)\in\argmin_{\mu,\nu}\gL(\mu,\nu), C\in\R,C'\in\R\}
\]

$(\supseteq)$
Let $(\mu^*,\nu^*)\in\argmin_{\mu,\nu}\gL(\mu,\nu)$. Then, 
\begin{align*}
    \frac{d^*(s,a)}{d^I(s,a)}=w^*_{\mu^*,\nu^*}(s,a)
    &=\exp\Big(\tfrac{1}{\alpha} \big( \E_{s'\sim T(\cdot|s,a)} [\gamma\nu(s')-\mu(s,s')]-\nu(s) \big) -1\Big),\\
    \frac{d^*(s,s')}{d^I(s,s')}=\bar{w}^*_{\mu^*,\nu^*}(s,s')
    &=\exp(r(s,s')+\mu^*(s,s')-1).
\end{align*}

Let $C\in\R$ and $C'\in\R$. Then, we derive the following equation:
\begin{align*}
    &\gL_\text{FD}(\mu^*+C,\nu^*+C')\\
    &=\gL_\text{FD}(\mu^*,\nu^*)\\
    &=(1-\gamma)\E_{s\sim p_0}[\nu^*(s)]+\log\E_{(s,a,s')\sim d^I}[\exp(\mu^*(s,s')+r(s,s'))]\\
    &\quad+\alpha\log\E_{(s,a)\sim d^I}\Big[\exp\Big(\tfrac{1}{\alpha} \big( \E_{s'\sim T(\cdot|s,a)}[\gamma\nu^*(s')-\mu^*(s,s')]-\nu^*(s) \big) \Big)\Big]\\
    &=(1-\gamma)\E_{s\sim p_0}[\nu^*(s)]+\log\E_{(s,a,s')\sim d^I}[\bar w^*_{\mu^*,\nu^*}(s,s')\exp(1)]\\
    &\quad+\alpha\log\E_{(s,a)\sim d^I}[w(s,a)\exp(1)]\\
    &=(1-\gamma)\E_{s\sim p_0}[\nu^*(s)]+\log\E_{(s,a,s')\sim d^*}[\exp(1)]+\alpha\log\E_{(s,a)\sim d^*}[\exp(1)]\\
    &=(1-\gamma)\E_{s\sim p_0}[\nu^*(s)]+1+\alpha\\
    &=(1-\gamma)\E_{s\sim p_0}[\nu^*(s)]+1+\alpha.
\end{align*}
Here, we apply Lemma~\ref{lem:fd_const_invariance} to derive the first equality. Because
\begin{align*}
    &\min_{\mu,\nu}\gL_\text{FD}(\mu,\nu)\\
    &=\min_{\mu,\nu}\gL(\mu,\nu)\\
    &=\gL(\mu^*,\nu^*)\\
    &=(1-\gamma)\E_{s\sim p_0}[\nu^*(s)]+\E_{(s,a,s')\sim d^I}[\exp(\mu^*(s,s')+r(s,s')-1)]\\
    &\quad+\alpha\E_{(s,a)\sim d^I}\Big[\exp\Big(\tfrac{1}{\alpha} \big( \E_{s'\sim T(\cdot|s,a)}[\gamma\nu^*(s')-\mu^*(s,s')]-\nu^*(s)\big)-1\Big)\Big]\\
    &=(1-\gamma)\E_{s\sim p_0}[\nu^*(s)]+\E_{(s,a,s')\sim d^I}[\bar{w}^*_{\mu^*,\nu^*}(s,s')]+\alpha\E_{(s,a)\sim d^I}[w^*_{\mu^*,\nu^*}(s,a)]\\
    &=(1-\gamma)\E_{s\sim p_0}[\nu^*(s)]+\E_{(s,a,s')\sim d^*}[1]+\alpha\E_{(s,a)\sim d^*}[1]\\
    &=(1-\gamma)\E_{s\sim p_0}[\nu^*(s)]+1+\alpha,
\end{align*}
we can conclude that $\gL_\text{FD}(\mu^*+C,\nu^*+C')=\min_{\mu,\nu}\gL_\text{FD}(\mu,\nu)$, i.e., $(\mu^*+C,\nu^*+C')\in\argmin_{\mu,\nu}\gL_\text{FD}(\mu,\nu)$. It means,
\[
\argmin_{\mu,\nu}\gL_\text{FD}(\mu,\nu)\supseteq\{\mu+C,\nu+C'|(\mu,\nu)\in\argmin_{\mu,\nu}\gL(\mu,\nu), C\in\R,C'\in\R\}
\]
\end{proof}

Remark that the proposed objective~\eqref{eq:dual_fenchel} is stable and the aforementioned proposition allows us to use self-normalized weighted importance sampling to extract policy. However, as we discussed in Section~\ref{subsec:pratical_algorithm}, using $\mu$ is main bottleneck for the overall optimization and optimizing $\tilde{\gL}(\nu)$ shows better performance in practice.

\clearpage
\section{Pseudocode of LobsDICE}
\label{app:pseudocode}

\subsection{Tabular LobsDICE}
\label{app:pseudocode_tabular}
For tabular MDPs, we first construct a maximum-likelihood estimation (MLE) MDP $\hat M = \langle S, A, \hat T, \hat p_0, \gamma \rangle$ using an offline dataset $D^I$.
Then, tabular LobsDICE solves \eqref{eq:lagrange_objective_double} on the MLE MDP $\hat M$.
In tabular case, note that $\widehat{d}^I(s,a)$, $\widehat{d}^I(s,s')$, and $\widehat{d}^E(s,s')$ are explicitly accessible by normalized counts of $(s,a)$ and $(s,s')$ in the datasets.
In addition, $\widehat{r}(s,s')=\frac{\widehat{d}^E(s,s')}{\widehat{d}^I(s,s')}$ is the log ratio between two empirical distributions.
Also, $\nu \in \R^{|S|}$ is represented as a $|S|$-dimensional vector, and $\mu \in \R^{|S|\times|S|}$ is represented as a $|S|^2$-dimensional vector, and we solve the following convex minimization problem for $(\mu, \nu)$:
\begin{align}
    \min_{\mu,\nu}\, &
    \textstyle J (\mu, \nu) 
    = (1-\gamma) \sum\limits_{s} \hat p_0(s) \nu(s) + \sum\limits_{s,s'} \widehat{d}^I(s,s') \Big[ \exp\big(\widehat r(s,s') + \mu(s,s')-1\big) \Big] \nonumber \\
    & \textstyle  + \alpha \sum\limits_{s,a} \widehat{d}^I(s,a) \Big[ \exp \Big( \tfrac{1}{\alpha} \sum\limits_{s'} \hat T(s'|s,a) \big( -\mu(s,s') + \gamma \nu(s') - \nu(s) \big) - 1 \Big) \Big]
    \label{eq:tabular_lobsdice_objective}
\end{align}
This process is summarized in Algorithm~\ref{alg:tabular_lobsdice}.
\begin{algorithm}[h!]
\caption{Tabular LobsDICE}
\centering
\begin{algorithmic}[1]
	\REQUIRE state-only demonstrations by experts $D^E = \{(s, s')_i\}_{i=1}^{N_E}$, 
	state-action demonstrations by some imperfect agents $D^I = \{(s,a,s')_i\}_{i=1}^{N_I}$,
	a learning rate $\eta$.
	\STATE Construct an MLE MDP $\hat M = \langle S, A, \hat T, \hat p_0, \gamma \rangle$ via normalized visitation counts of $D^I$.
	\STATE Construct $\widehat{d}^I(s,a)$, $\widehat{d}^I(s,s')$, and $\widehat{d}^E(s,s')$ via normalized visitation counts of $D^I$ and $D^E$.
	\STATE $\widehat{r}(s,s') \leftarrow \log \frac{\widehat{d}^E(s,s')}{\widehat{d}^I(s,s')}$ for all $s,s'$.
	\STATE Randomly initialize $\nu$ and $\mu$.
	\WHILE{$(\nu, \mu)$ is not converged}
	    \STATE $(\mu, \nu) = (\mu, \nu) - \eta \nabla_{\mu,\nu} J(\mu, \nu)$ ~~~ (Eq.~\eqref{eq:tabular_lobsdice_objective})
	\ENDWHILE
	\STATE $w^*(s,a) = \exp\Big(\tfrac{1}{\alpha} \big(\sum_{s'} \hat T(s'|s,a) (-\mu(s,s') + \gamma \nu(s') - \nu(s) ) \big) -1\Big) $ for all $s,a$. ~ (Eq.~\eqref{eq:closed_form_solution_wsa_wss1})
	\STATE $\pi^*(a|s) \leftarrow \frac{\widehat{d}^I(s,a) w^*(s,a)}{ \sum_{a'} \widehat{d}^I(s,a') w^*(s,a') }$ for all $s,a$.
	\ENSURE The imitation policy $\pi^*$.
\end{algorithmic}
\label{alg:tabular_lobsdice}
\end{algorithm}

\subsection{Practical LobsDICE (with function approximation)}
\label{app:pseudocode_continuous}
To deal with continuous or large MDPs, we represent our optimization variable $\nu$ and discriminator $c$ as neural networks, parameterized by $\theta$ and $\phi$ respectively: $\nu_\theta: S \rightarrow \R$ is an MLP that takes a state as an input and outputs a scalar value, and $c_\phi: S \times S \rightarrow [0, 1]$ is defined similarly.
For the policy $\pi_\psi$, we use a tanh-squashed Gaussian policy, where the parameters of a Gaussian distribution (i.e. mean and covariance) are output by the neural network.

The parameters of $c_\phi$ are trained by:
\begin{align}
    \max_{\phi} J_c(\phi) := 
    \E_{\substack{\mathrm{batch}(\bar d^E) \sim \bar d^E \\ \mathrm{batch}(\bar d^I) \sim \bar d^I}}
    \Big[ \bE_{(s,s') \sim \mathrm{batch}(\bar{d}^E)}[\log c_\phi(s,s')] 
    + \bE_{(s,s') \sim \mathrm{batch}(\bar{d}^I)}[\log (1-c_\phi(s,s'))] \Big],
    \label{eq:loss_J_c}
\end{align}
which is analogous to \eqref{eq:discriminator_training}.
The parameters of $\nu_\theta$ are trained by:
\begin{align}
    &\min_{\theta} J_\nu(\theta) := 
    \E_{\substack{ \mathrm{batch}(d^I) \sim d^I \\ \mathrm{batch}(p_0) \sim \hat p_0 }} \Big[
    (1-\gamma) \bE_{s \sim \mathrm{batch}(p_0)}[\nu_\theta(s)] 
    \label{eq:loss_J_nu}
    \\ &\hspace{80pt} 
    + (1 + \alpha) \log \bE_{(s,a,s') \sim \mathrm{batch}(d^I)} [\exp(\tfrac{1}{1+\alpha} \hat A_{\theta}(s,a,s'))] \nonumber
    \Big],
\end{align}
where $\hat A_{\theta}(s,a,s') = r_\phi(s,s') + \gamma \nu_\theta(s') - \nu_\theta(s)$ and $r_\phi(s,s') = -\log\Big( \tfrac{1}{c_\phi(s,s')} - 1 \Big)$, and this is analogous to \eqref{eq:main_fenchel_objective}.
Due to the logarithm outside the expectation in the second term, the mini-batch approximation would introduce additional bias in gradient estimation, but we found that using the biased gradient estimate worked well in practice with moderately large batch size (e.g. 256).
Finally, the parameters of $\pi_\psi$ are trained by:
\begin{align}
    \max_{\psi} J_\pi(\psi)
    :=
    \E_{\mathrm{batch}(d^I) \sim d^I} \left[
    \frac
    {\bE_{(s,a,s') \sim \mathrm{batch}(d^I)} \Big[ \exp \big(\tfrac{1}{1+\alpha} \hat A_{\theta}(s,a,s') \big) \log \pi_\psi(a|s) \Big] }
    {\bE_{(s,a,s') \sim \mathrm{batch}(d^I)} \Big[ \exp \big(\tfrac{1}{1+\alpha} \hat A_{\theta}(s,a,s') \big) \Big]}
    \right],
    \label{eq:loss_J_pi}
\end{align}
which is analogous to \eqref{eq:self_normalized_weighted_bc}.
Here, the self-normalized importance sampling may introduce an additional bias with mini-batch approximations, but it is well known that the self-normalized importance sampling provides a consistent estimator~\citep{owen2013snis}. Also, it worked well in practice with moderately large batch size (e.g. 256) in our experiments.
We optimize the parameters $(\phi, \theta, \psi)$ jointly in practice.
We are not using a target network at all.
The overall training process is summarized in Algorithm~\ref{alg:practical_lobsdice}.

\begin{algorithm}[h!]
\caption{LobsDICE (with function approximation)}
\centering
\begin{algorithmic}[1]
	\REQUIRE state-only demonstrations by experts $D^E = \{(s, s')_i\}_{i=1}^{N_E}$, 
	state-action demonstrations by some imperfect agents $D^I = \{(s,a,s')_i\}_{i=1}^{N_I}$,
	a learning rate $\eta$.
	\STATE Initialize parameter vectors $\phi, \theta, \psi$.
	\FOR{each gradient step}
	    \STATE Sample mini-batches from $D^E$ and $D^I$.
	    \STATE Compute gradients and perform SGD update:
	    \STATE ~~~~$\phi \leftarrow \phi + \eta \nabla_\phi J_c(\phi)$ 
	    \hspace{34pt} (Eq.~\eqref{eq:loss_J_c})
	    \STATE ~~~~$\theta \leftarrow \theta - \eta \nabla_\theta J_{\nu}(\theta)$ \hspace{36pt} (Eq.~\eqref{eq:loss_J_nu})
	    \STATE ~~~~$\psi \leftarrow \psi + \eta \nabla_\psi J_\pi(\psi)$ ~~
	    \hspace{22pt} (Eq.~\eqref{eq:loss_J_pi})
	\ENDFOR
	\ENSURE The imitation policy $\pi_{\psi}$.
\end{algorithmic}
\label{alg:practical_lobsdice}
\end{algorithm}

\clearpage

\section{Additional Experiments}
\label{app:additinoal_experiments}
\subsection{Subsampled expert demonstrations}
Existing works on imitation learning~\citep{ho2016gail} conduct experiments where the expert demonstrations are subsampled.
The following figure presents the result when given the subsampled expert demonstrations. LobsDICE still significantly outperforms the baseline algorithms even with the subsampled expert demonstrations\footnote{Recently, \citeauthor{li2022rethinking} reported that ValueDICE~\citep{kostrikov2020valuedice} (without target network) significantly underperforms DAC~\citep{kostrikov2019discriminator} (with target network) in the subsampled expert trajectories setting due to divergence issue and pointed to the absence of the target network in ValueDICE as a reason for performance degradation.
However, we didn't observe any divergence issue due to the absence of the target network, and LobsDICE still performs well even when the trajectories are subsampled.}.
\begin{figure}[h]
\vspace{-0.4cm}
\includegraphics[width=.95\linewidth]{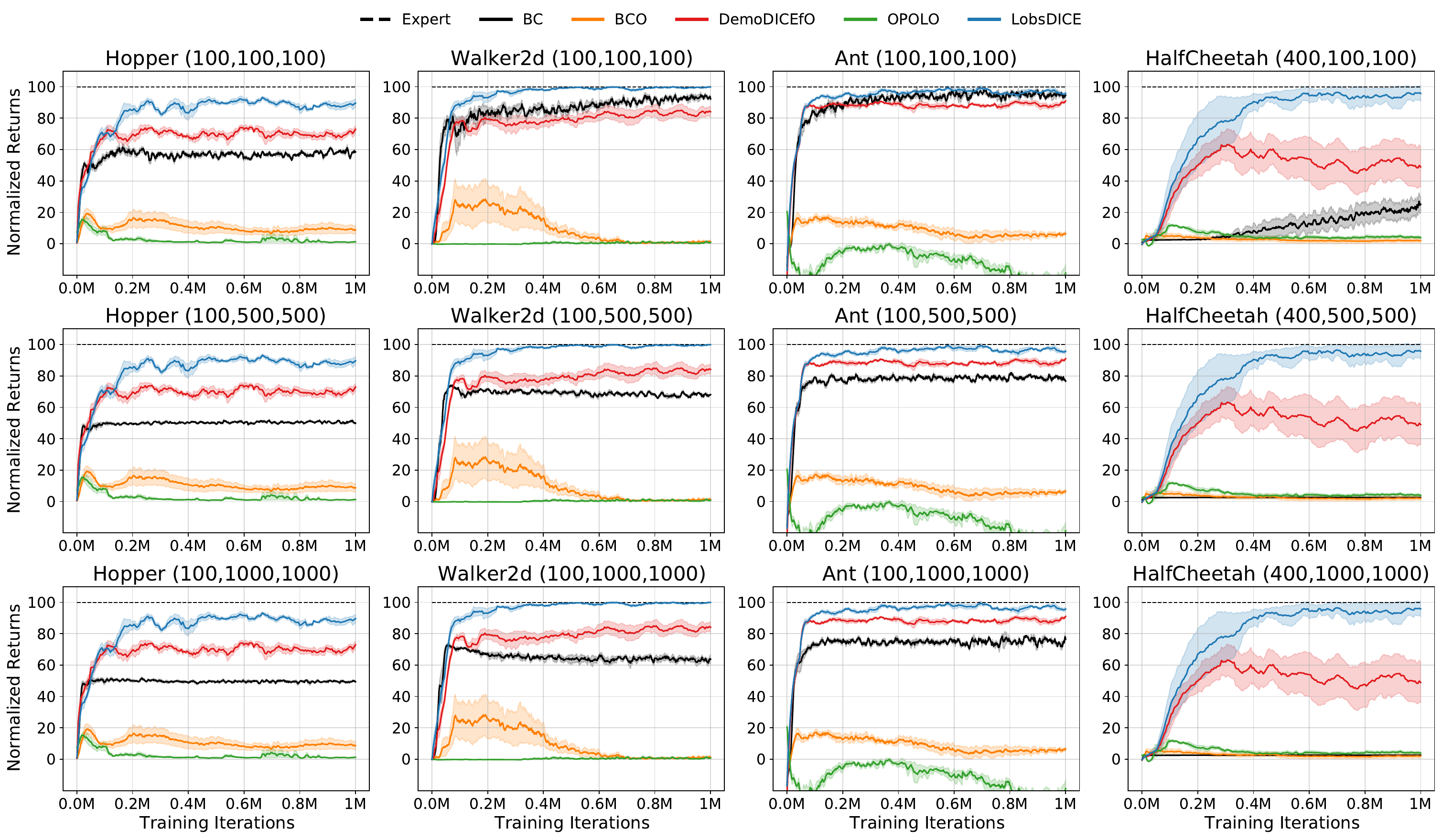}
\vspace{-0.3cm}
\caption{Performance of LobsDICE and baseline algorithms on various MuJoCo control tasks.
We build state-only expert demonstrations using 50 subsampled trajectories from \texttt{expert-v2} (subsampling rate is 20).
}
\end{figure}

\vspace{-0.5cm}
\subsection{Error of inverse dynamics model in MuJoCo domains}
\begin{figure}[h]
\vspace{-0.35cm}
\includegraphics[width=.95\linewidth]{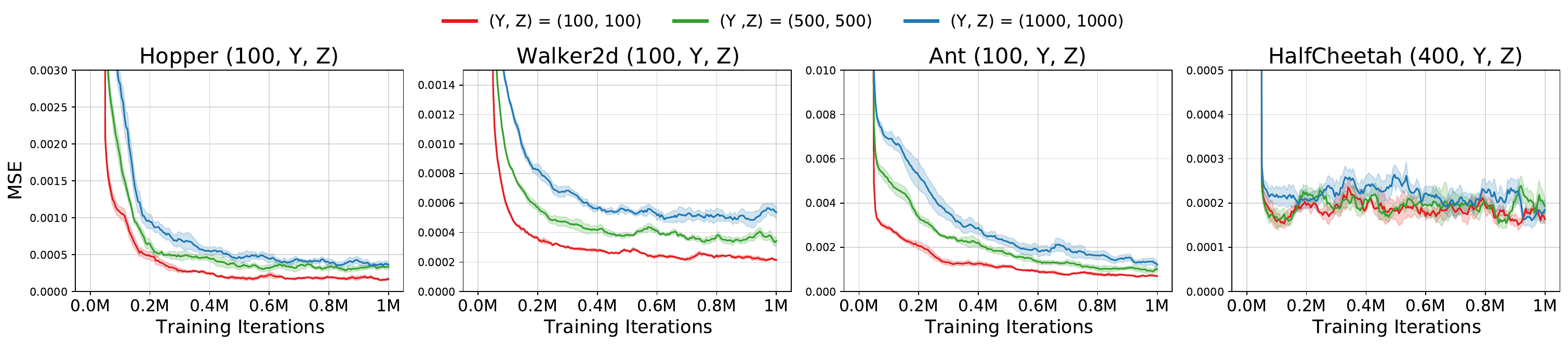}
\vspace{-0.3cm}
\caption{Mean squared error of IDM for expert demonstrations $D^E$. For each task $(Y,Z)$, we construct imperfect demonstrations using $100$ ($400$ for HalfCheetah),  $Y$, and $Z$ trajectories from \texttt{expert-v2}, \texttt{medium-v2}, and \texttt{random-v2}, respectively.}
\label{fig:idm_error}
\end{figure}

In principle, IDM trained with arbitrary demonstrations should be able to predict the expert's missing actions accurately in deterministic MDPs.
However, in Section~\ref{subsec:experiments_mujoco}, we observed that the empirical performance of BCO and DemoDICEfO tends to degrade as the number of imperfect demonstrations increases (see Figure~\ref{fig:mujoco}).
This is due to the use of function approximation for IDM: given that the expressive power of a function approximator is limited, the more data unrelated to the expert demonstrations (i.e. \texttt{medium-v2} and \texttt{random-v2}) is used for training, the more the prediction accuracy for the expert data would be adversely affected.
Figure~\ref{fig:idm_error} shows that the mean squared error of IDM for expert demonstrations $D^E$ increases as the number of non-expert demonstrations increases.

\begin{figure*}[h]
\centering
\includegraphics[width=\linewidth]{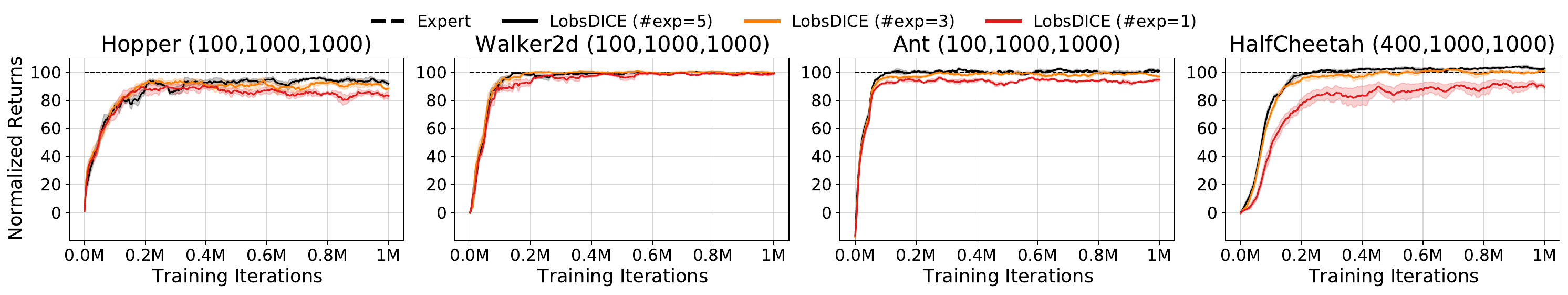}
\caption{
\revised{
Performance of LobsDICE on various MuJoCo control tasks.
We build state-only expert demonstrations using 1, 3, and 5 trajectories from \texttt{expert-v2}.
For each task $(X,Y,Z)$ we construct imperfect demonstrations using
$X$, $Y$, and $Z$ trajectories from \texttt{expert-v2}, \texttt{medium-v2}, and \texttt{random-v2}, respectively.
We plot the mean and the standard errors (shaded area) of the normalized scores over five random seeds.}
}
\label{fig:mujoco_fewer_expert}
\end{figure*}

\revised{
\subsection{Fewer expert demonstrations}
\label{app:fewer_expert_demo}
In this section, we provide additional experiments with fewer expert demonstrations.
Specifically, we tested LobsDICE when the number of state-only expert demonstrations is 5, 3, and 1, respectively.
The Figure~\ref{fig:mujoco_fewer_expert} shows that LobsDICE (\#exp=3) and LobsDICE (\#exp=1) still perform well.
}

\clearpage
\section{Experimental Details}
\label{app:experiment_detail}
\subsection{Experimental details for random MDPs}
\label{app:experiment_detail_random_mdp}
\paragraph{Random MDP generation}
For MDP $M=\langle S,A,T,R,\gamma,p_0\rangle$, we first set $|S|=20$, $|A|=4$, $\gamma=0.95$, $p_0(s)=1$ for a fixed $s=s_0$. For each $(s,a)$, we samples four different state $\{s_1',s_2',s_3',s_4'\}$. Then, we set the transition probability $(T(s_1'|s,a),T(s_2'|s,a),T(s_3'|s,a),T(s_4'|s,a))=(1-\beta) X+\beta Y$, where $X\sim\text{Categorical}(0.25,0.25,0.25,0.25)$ and $Y\sim\text{Dir}(1,1,1,1)$.
$\beta\in[0,1]$ is the hyperparameter to control transition stochasticity: when $\beta=0$, the transition probability becomes one-hot vector, i.e., deterministic MDP. In contrast, when $\beta=1$, the transition of MDP becomes stochastic. Finally, the reward of 1 is only given to a state that minimizes the optimal value at initial state $s_0$; other states have zero rewards. 

\paragraph{Offline dataset generation}
For each random MDP $M$, we generate state-only expert demonstrations by executing a (softmax) expert policy and state-action imperfect demonstrations by a uniform random policy. We perform experiments for a varying number of expert demonstrations $N_E \in \{10, 100, 1000, 1000\}$ and imperfect demonstrations $N_I \in \{1, 3, 10, 30, 100, 300, 1000, 3000, 10000\}$.

\paragraph{Hyperparameters}
We compare our tabular LobsDICE with BC, BCO, DemoDICEfO, and OPOLO. For KL-regularization hyperparameters of DemoDICEfO and LobsDICE, we use $\alpha=0.1$.

\subsection{Experimental details for MuJoCo control tasks}
\label{app:experiment_detail_mujoco}

\begin{table}[!h]
\centering
\begin{adjustbox}{center} 
\begin{tabular}{@{}l|cccccc@{}}\toprule
Hyperparameters                & BC     & BCO       & DemoDICEfO    & LobsDICE    \\ \midrule\midrule
$\gamma$ (discount factor)     & 0.99   & 0.99      & 0.99          & 0.99 \\
$\alpha$ (regularization coefficient)   & -  & -    & 0.1           & 0.1\\\midrule
learning rate (actor)  & $3\times 10^{-4}$ & $3\times 10^{-4}$ & $3\times 10^{-4}$& $3\times 10^{-4}$\\
network size  (actor)  & [256, 256]     & [256, 256]    & [256, 256] & [256, 256]\\
learning rate (critic) & -          & - & $3\times 10^{-4}$& $3\times 10^{-4}$\\
network size  (critic) & -          &  -        &  [256,256] & [256, 256]\\
learning rate (discriminator)   & -     &  -    &  $3\times 10^{-4}$& $3\times 10^{-4}$\\
network size  (discriminator)   & -     &  -    &  [256,256]& [256, 256]\\
learning rate (inverse dynamics)   & -    &  $3\times 10^{-4}$& $3\times 10^{-4}$ &  -    \\
network size  (inverse dynamics)   & -     &  [256,256]& [256, 256] &  -   \\ \midrule
gradient L2-norm coefficient (critic)  & -      &  - &  $1\times 10^{-4}$& $1\times 10^{-4}$\\
gradient penalty coefficient (discriminator)  & -      & -  & 0.1  & 0.1 \\
\midrule
batch size                 & 512 & 512 & 512 & 512 \\
\# of expert trajectories  & 5   & 5   & 5   & 5 \\
\# of training iterations & 1,000,000 & 1,000,000 & 1,000,000 & 1,000,000 \\
\bottomrule
\end{tabular}
\end{adjustbox}
\caption{Configurations of hyperparameters used in our experimental results.}
\label{hyperparams}
\end{table}

For fair comparison, we use the same learning rate to train actors of BC, BCO, DemoDICEfO, and LobsDICE.
We implement our network architectures for BC, BCO, DemoDICEfO, and LobsDICE based on the implementation of OptiDICE\footnote{\scriptsize\url{https://github.com/secury/optidice}}.
For OPOLO, we use its official implmentation\footnote{\scriptsize\url{https://github.com/illidanlab/opolo-code}}.
We have tried hyperparameter tuning for OPOLO but never obtained a successful learning curve, showing numerical instability due to using out-of-distribution action values.
Vanilla \emph{online} off-policy algorithms commonly fail in the \emph{offline} learning setting.
Therefore, we report the results of official implementation without any modification to network architectures or hyperparameters.
For stable discriminator learning, we use gradient penalty regularization on the $r(s,a)$ and $r(s,s')$ functions, which was proposed in \cite{gulrajani2017improvedgan} to enforce 1-Lipschitz constraint. 
To stabilize critic training, we add gradient L2-norm to the critic loss for the regularization.
In addition, we use every state $s$ in $D^I$ to define $D_0$, i.e., $D_0=\{s|(s,a,s')\in D^I\}$, following ValueDICE~\citep{kostrikov2020valuedice}.
Detailed hyperparameter configurations used for our main experiments are summarized in Table \ref{hyperparams}.

\paragraph{Evaluation metric}
For each environment, the normalized score is measured by $100\times\frac{\texttt{score}-\texttt{random score}}{\texttt{expert score}-\texttt{random score}}$, where the \texttt{expert score} and \texttt{random score} are average returns of trajectories in \texttt{expert-v2} and \texttt{random-v2}, respectively.

\begin{figure*}[h]
\centering
\includegraphics[width=\linewidth]{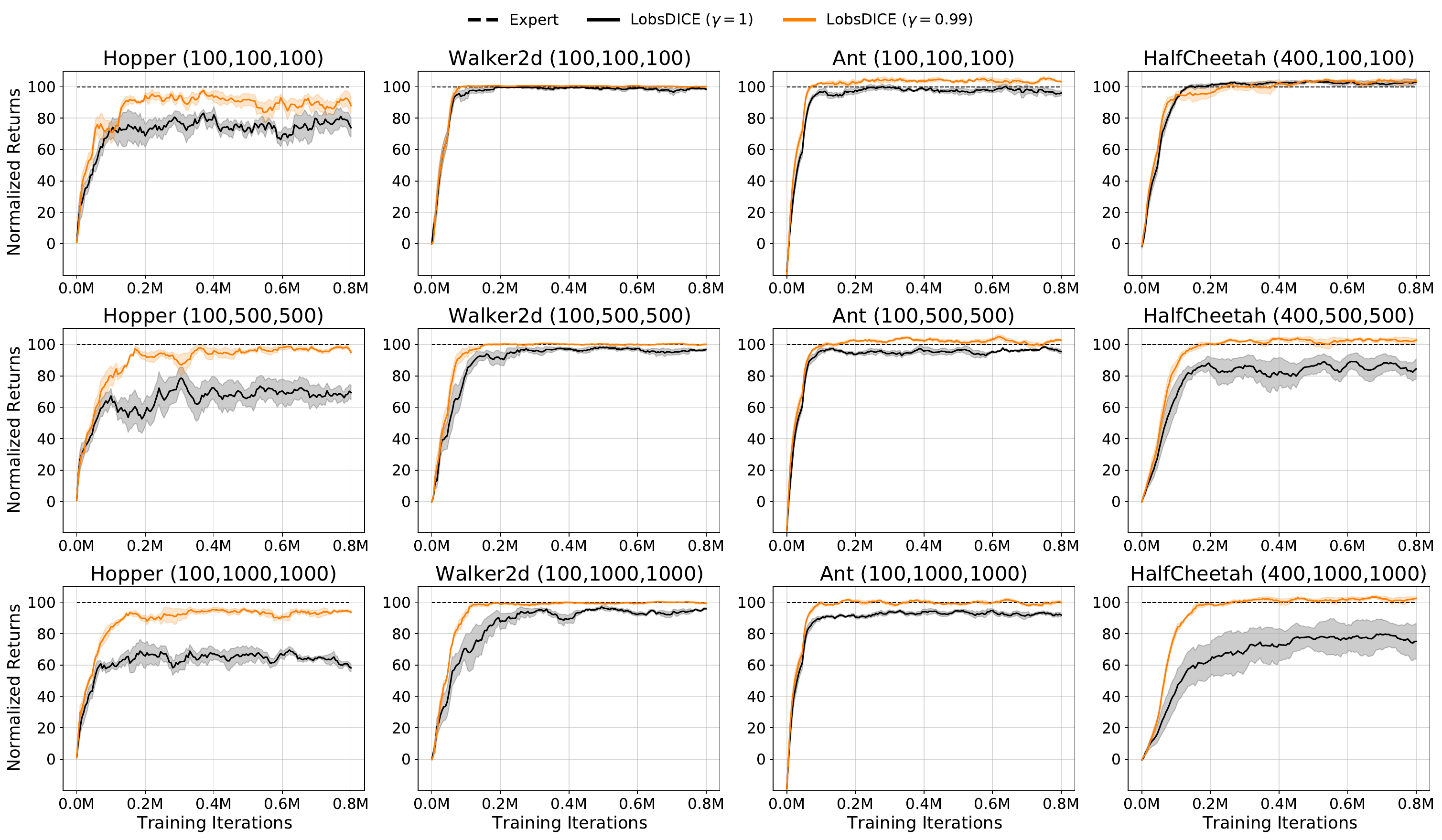}
\caption{
\revised{
Performance of LobsDICE ($\gamma=0.99$) and LobsDICE ($\gamma=1$) on various MuJoCo control tasks.
We build state-only expert demonstrations using 5 trajectories from \texttt{expert-v2}. 
For each task $(X,Y,Z)$ we construct imperfect demonstrations using
$X$, $Y$, and $Z$ trajectories from \texttt{expert-v2}, \texttt{medium-v2}, and \texttt{random-v2}, respectively.
We plot the mean and the standard errors (shaded area) of the normalized scores over five random seeds.}
}
\label{fig:mujoco_gamma_1}
\end{figure*}

\revised{
\section{Generalization to $\gamma=1$}
\label{app:gamma_1}
In this section, we generalize LobsDICE for discounted MDPs ($\gamma<1$) to undiscounted MDPs ($\gamma=1$).
Suppose that the stationary distributions $d$ and $\bar{d}$ satisfy the Bellman flow constraints~\eqref{eq:bellman_flow_constraint} and marginalization constraints~\eqref{eq:marginalization_constraint}, respectively. 
When $\gamma=1$, for any constant $c\ge0$, $cd$ and $c\bar{d}$ also satisfy the corresponding constraints.\\
It means that the original constrained optimization problem (\ref{eq:main_objective}-\ref{eq:marginalization_constraint}) for the stationary distributions $d$ and $\bar{d}$ is an ill-posed problem.
We tackle this ill-posedness by adding additional normalization constraint $\sum_{s,a}d(s,a)=1$ to (\ref{eq:main_objective}-\ref{eq:marginalization_constraint}):
\begin{align*}
    \max_{\blue{d},\redb{d}\ge0}~ 
    &-\KL(\redb{d}(s,s') \| \bar{d}^E(s,s') ) - \alpha\KL(\blue{d}(s,a) \| d^I(s,a)) \\
    \text{s.t.}~
    &\textstyle\sum\limits_{a'} \blue{d}(s',a') = (1-\gamma)p_0(s') + \gamma \sum\limits_{s,a} \blue{d}(s,a) T(s'|s,a) \quad\forall s',\\
    &\textstyle \sum\limits_{a}\blue{d}(s,a) T(s'|s,a) = \redb{d}(s,s')\quad\forall s,s',\\
    &\textstyle \sum\limits_{s,a}\blue{d}(s,a) = 1.
\end{align*}
Using a derivation similar to that from \eqref{eq:lagrangian} to \eqref{eq:maximin_lagrangian_detail}, we obtain the following min-max optimization for $\blue{w},\redb{w},\mu, \nu$, and $\lambda$:
\begin{align}
&\min_{\mu,\nu,\lambda}\max_{\blue{w},\redb{w}\ge0}\gL(\blue{w},\redb{w},\mu,\nu,\lambda)\nonumber\\
&:=\gL(\blue{w},\redb{w},\mu,\nu)+\lambda(1-\E_{(s,a)\sim d^I}[\blue{w}(s,a)])\nonumber\\
&=(1 - \gamma) \E_{s_0 \sim p_0} [\nu(s_0)] + \E_{(s,s') \sim \bar{d}^I} \big[ \redb{w}(s,s') \big( r(s,s') + \mu(s,s') - \log\redb{w}(s,s') \big)\big] \nonumber \\
& \hspace{30pt} + \E_{(s,a) \sim d^I} \big[\blue{w}(s,a) \big( \underbrace{e_{\mu,\nu}(s,a)  -\lambda}_{:=e_{\mu,\nu,\lambda}(s,a)} - \alpha\log\blue{w}(s,a)\big)\big] + \lambda,
\label{eq:lagrangian_generalized}
\end{align}
where $\lambda\in\R$ is the Lagrange multiplier for the normalization constraint $\sum_{s,a}d(s,a)=1$. 
Similar to Proposition~\ref{prop:closed_form_solution}, we can derive a closed-form solution to the inner maximization of~\eqref{eq:lagrangian_generalized}:
\begin{align*}
    \blue{w}_{\mu,\nu,\lambda}(s,a)&=\exp\big(\tfrac{1}{\alpha} e_{\mu,\nu,\lambda}(s,a) -1\big) 
    \emph{ ~and~ }
    \redb{w}_{\mu}(s,s')=\exp(r(s,s')+\mu(s,s')-1).
\end{align*}
Then, we can reduce the nested min-max optimization of~\eqref{eq:lagrangian_generalized} to a single minimization by plugging the closed-form solution $(\blue{w}_{\mu,\nu,\lambda}, \redb{w}_{\mu})$ into $\gL(\blue{w},\redb{w},\mu,\nu,\lambda)$:
\begin{align}
    \min_{\mu,\nu,\lambda}\, & \gL(\blue{w}_{\mu,\nu,\lambda},\redb{w}_\mu,\mu,\nu,\lambda)
    = (1-\gamma)\E_{s \sim p_0}[\nu(s)] \nonumber \\
    & + \E_{(s,s') \sim \bar{d}^I} \big[ \exp\big(r(s,s')+\mu(s,s')-1\big) \big] + \alpha\E_{(s,a) \sim d^I}\big[ \exp\big(\tfrac{1}{\alpha} e_{\mu,\nu,\lambda}(s,a) -1\big)\big]+\lambda. \nonumber
\end{align}
In practice, we estimate $\gL(\blue{w}_{\mu,\nu,\lambda},\redb{w}_\mu,\mu,\nu,\lambda))$ using samples from distribution $d^I$. To derive a sample-based objective, we use analogous derivation in Section~\ref{subsec:pratical_algorithm}.
Sample-based objective can be represented as
\begin{align}
    \min_{\mu,\nu,\lambda}\, & \widehat\gL(\mu,\nu,\lambda)
    = (1-\gamma)\bE_{s\in D_0}[\nu(s)] \label{eq:practical_lagrangian_triple} \\
    & + \bE_{x\in D^I} \big[ \exp\big(r(s,s')+\mu(s,s')-1\big) + \alpha\exp\big(\tfrac{1}{\alpha} \hat{e}_{\mu,\nu,\lambda}(s,a,s') -1\big)\big]+\lambda,\nonumber
\end{align}
where $\hat{e}_{\mu,\nu,\lambda}(s,a,s')=-\mu(s,s')+\gamma\nu(s')-\nu(s)-\lambda$.
Then, the closed-form solution to the minimization~\eqref{eq:practical_lagrangian_triple} with respect to $\mu$ is:
\begin{equation}
    \mu_{\nu,\lambda}(s,s')=\frac{1}{1+\alpha}\big(-\alpha r(s,s')+\gamma\nu(s')-\nu(s)-\lambda\big)
\end{equation}
Using the above solution, we obtain the following minimization problem:
\begin{align}
    \min_{\widehat\nu,\widehat\lambda}\, & \widehat\gL(\widehat\nu,\widehat\lambda)
    = (1-\gamma)\bE_{s\in D_0}[\nu(s)]+\widehat\lambda
    + (1+\alpha) \bE_{x\in D^I} \Big[ \exp\big(\tfrac{1}{1+\alpha}\widehat{A}_{\widehat\nu,\widehat\lambda}(s,a,s')-1\big)\Big], \label{eq:practical_lagrangian_double}
\end{align}
where $\widehat{A}_{\nu,\lambda}(s,a,s'):=r(s,s')+\gamma\nu(s')-\nu(s)-\lambda$. 
Finally, similar to Proposition~\ref{prop:fencheldual}, we can obtain the following objective with the same minimum as \eqref{eq:practical_lagrangian_double}:
\begin{align}
    \min_{\nu,\lambda}\widetilde \gL (\widetilde\nu,\widetilde\lambda)
    & = (1-\gamma) \bE_{s \in D_0}[\widetilde\nu(s)] + \widetilde\lambda + (1+\alpha) \log \bE_{x \in D^I} \Big[\exp\Big(\tfrac{1}{1+\alpha} \widehat A_{\widetilde\nu,\widetilde\lambda}(s,a,s')\Big)\Big]\nonumber\\
    & = (1-\gamma) \bE_{s \in D_0}[\widetilde\nu(s)] + (1+\alpha) \log \bE_{x \in D^I} \Big[\exp\Big(\tfrac{1}{1+\alpha} \widehat A_{\widetilde\nu}(s,a,s')\Big)\Big]=\min_{\nu}\widetilde \gL (\widetilde\nu).\nonumber
\end{align}
Interestingly, the resulting objective is the same as the original objective $\widetilde \gL (\widetilde\nu)$ of \eqref{eq:main_fenchel_objective}.
Based on this theoretical result, we compare LobsDICE for $\gamma=1$, denoted by LobsDICE ($\gamma=1$), with LobsDICE for $\gamma=0.99$, denoted by LobsDICE ($\gamma=0.99$).
In Figure~\ref{fig:mujoco_gamma_1}, LobsDICE ($\gamma=1$) shows good performance, but LobsDICE ($\gamma=0.99$) is more stable and achieves better performance in practice.}

\section{Computation Resources}
\label{app:resource}
We used 10 servers equipped with the following specification:
\begin{itemize}
    \item CPU: Intel(R) Core(TM) i7-9700K CPU @ 3.60GHz.
    \item Memory: 32 GB.
    \item GPU: TITAN V.
\end{itemize}

\section{Limitation}
\label{app:limitation}
Although we demonstrated that LobsDICE successfully recovered the expert's behavior in the experiments, it requires the assumption that support of $D^I$ covers $D^E$.
Relaxing this assumption remains as a future work.

\section{License}
\label{app:license}
D4RL~\citep{fu2020d4rl} is licensed under the Apache 2.0

\section{Social Impact}
\label{app:social_impact}
This work contributes to an algorithmic foundation for imitation learning from observation in an offline setting.
Given that state-only expert demonstrations are much easier to be collected than the action-labeled expert demonstrations and that the imperfect demonstrations of arbitrary optimality is also easy to be collected, our method has a potential to be widely adopted in many real-world applications.
On the other hand, this work could adversely affect employment by contributing to the automation of tasks having done by human experts (e.g. factory automation, autonomous driving).

\end{document}